\documentclass[manuscript,screen]{acmart}

\usepackage{amsmath}
\usepackage{amsfonts}
\usepackage{algorithm}
\usepackage{algorithmic}
\usepackage{array}
\usepackage{booktabs}
\usepackage{graphicx}
\usepackage{multirow}
\usepackage{tabularx}

\setcopyright{none}
\acmJournal{CSUR}
\acmYear{2026}
\acmDOI{}
\newcolumntype{Y}{>{\centering\arraybackslash}X}
\newcolumntype{L}{>{\raggedright\arraybackslash}X}

\title{When Agentic AI Meets Integrated Sensing and Communication}

\author{Kai Li}
\affiliation{%
    \institution{Interdisciplinary Centre for Security, Reliability and Trust (SnT), University of Luxembourg}
    \city{Luxembourg}
    \country{Luxembourg}
}
\email{kaili@ieee.org}

\author{Conggai Li}
\affiliation{%
    \institution{Technology, CSIRO}
    \city{Sydney, NSW}
    \country{Australia}
}
\email{conggai.li@csiro.au}

\author{Sarah Ali Siddiqui}
\affiliation{%
    \institution{Technology, CSIRO}
    \city{Sydney, NSW}
    \country{Australia}
}
\email{sarahalisiddiqui@yahoo.com}

\author{Syed Sohail Ahmed}
\affiliation{%
    \institution{College of Computer, Qassim University}
    \city{Buraydah}
    \country{Saudi Arabia}
}
\email{sa.ahmed@qu.edu.sa}

\author{Xin Yuan}
\affiliation{%
    \institution{Technology, CSIRO}
    \city{Sydney, NSW}
    \country{Australia}
}
\email{xin.yuan@csiro.au}

\author{Shenghong Li}
\affiliation{%
    \institution{Technology, CSIRO}
    \city{Sydney, NSW}
    \country{Australia}
}
\email{shenghong.li@csiro.au}

\author{Wei Ni}
\affiliation{%
    \institution{Edith Cowan University, School of Engineering}
    \city{Perth, WA}
    \country{Australia}
}
\email{wei.ni@ieee.org}

\begin{document}






\begin{abstract}
Agentic artificial intelligence (AI) is transforming Integrated Sensing and Communication (ISAC) from a function-oriented physical-layer technology into a goal-driven, closed-loop intelligent system, a paradigm we term AISAC. Existing work on learning-based sensing, resource allocation, reconfigurable intelligent surfaces (RIS), edge intelligence, multi-agent coordination, and resilient networking has developed largely in isolation. This survey unifies the literature within a six-stage closed-loop framework comprising observation, contextualization, reasoning and prediction, planning and orchestration, execution and collaboration, and feedback and resilience. It also introduces five levels of agentic maturity, ranging from physical-layer primitives to fully closed-loop agentic ISAC. We use this framework to review advances in multimodal intelligence, large language models, reinforcement learning, federated learning, RIS-assisted control, Unmanned Aerial Vehicle (UAV) and vehicular networks, and AI-native network management, and analyze privacy, security, resilience, and sustainability as cross-cutting requirements of the full perception-reasoning-action loop. An audit of representative studies against nine agentic-specific evaluation criteria shows that no system reports more than one or two of them, exposing a gap between claimed and demonstrated agentic maturity. We identify open challenges in physical-to-semantic grounding, predictive world models, real-time agent-PHY interaction, safe tool use, heterogeneous multi-agent collaboration, benchmarking, and resource-efficient autonomy.
\end{abstract}

\maketitle

\section{Introduction}
\label{sec:intro}
\subsection{Background and Motivation}
Agentic artificial intelligence (AI) is a transformative paradigm for autonomous, target-driven intelligence~\cite{abdisarabshali2026large}. By integrating reasoning, planning, memory, and multi-agent collaboration, agentic AI enables wireless systems to act adaptively and cooperatively with minimal human supervision~\cite{lu2026agentic}.
There is the potential to combine agentic AI with Integrated Sensing and Communication (ISAC) to couple environmental perception, real-time information exchange, and distributed intelligence at the network edge. This supports latency-critical, context-aware Augmented Reality (AR)/Virtual Reality (VR)/Mixed Reality (MR) applications.

Agentic AI allows distributed edge devices to coordinate sensing and communication for latency-sensitive, mission-critical tasks~\cite{wu2026edge}. In particular, ISAC transmitters and receivers can be controlled by AI agents. The frequency, bandwidth, power control, and beamforming can be implemented by agentic AI. The radar signal processing can be executed by agents trained to infer different targets and recognize different radio fingerprints of objects~\cite{cai2026llm}. Through interactions between communication agents and sensing agents, effective ISAC strategies can be employed.

This nexus of agentic AI and ISAC is referred to \textit{AISAC, a new networking paradigm, in which autonomous AI agents, capable of perception, reasoning, planning, and decision-making, are embedded within ISAC systems to jointly orchestrate sensing and communication resources toward intent-driven, mission-level objectives.} AISAC is driven by a growing demand for federated intelligence and context-aware services,
with an aim to jointly optimize spectrum utilization, energy efficiency, and decision-making latency. It leverages edge computing to process sensed data locally~\cite{hou2026game, chen2026goodput}, while using communication links for cooperative exchange and distributed learning. 

This same nexus introduces new challenges. AISAC combines the vulnerabilities of agentic AI and ISAC into a single, multifaceted attack surface.
Conventional centralized learning collects raw data in the cloud. Agentic AI instead trains local models on private data and shares only model parameters or gradients~\cite{lin2026hsplitlora}. In AISAC, though, the integration of sensing and communication opens privacy leakage channels beyond traditional threats such as model inversion and membership inference~\cite{ni2026privacy, zhang2026agent}. 
AISAC typically shares spectrum between sensing and communication, which allows adversaries to intercept and analyze transmitted model updates or sensing signals. A distinctive threat is privacy leakage from sensing-computation correlation. ISAC tightly couples sensing acquisition with computational inference on the same device. This allows adversaries to exploit correlations between sensing signals, such as radar echoes or wireless reflections, and model outputs, such as predictions, to infer sensitive environmental states.  


Beyond privacy, AISAC also faces substantial resilience challenges. Malicious manipulation and unintended disruption both threaten its distributed, cooperative architecture, which is vulnerable to adversarial actions that degrade learning accuracy, compromise sensing integrity, and disrupt communication efficiency. Because sensing and communication are tightly coupled, poisoned updates can propagate through shared sensing features and communication signals, not just model parameters. This makes detection significantly harder. A compromised vehicle or Unmanned Aerial Vehicle (UAV), for instance, may inject falsified sensing data that biases the collaborative model toward incorrect environmental interpretations, triggering cascading errors across the entire ISAC system.


Cross-domain adversarial threats represent a further, newer frontier in AISAC security. ISAC runs signal processing, communication optimization, and learning in parallel, which gives adversaries room to design multi-layer manipulations. They might perturb sensing waveforms to influence feature extraction, or inject malicious gradients that alter waveform generation policies. Such cross-functional interference amplifies the impact of an attack. It also renders conventional, function-isolated defenses inadequate.

\subsection{Contribution}
Existing ISAC research has largely focused on jointly optimizing sensing and communication efficiency. AISAC goes further. It introduces a deeper level of integration, in which sensing, communication, computation, and autonomous decision-making become interdependent and continuously co-adaptive. Intelligent agents act as autonomous cognitive components that orchestrate sensing strategies, communication policies, and distributed inference.
Yet, the cross-domain coupling unique to AISAC creates a new class of threats. 
The literature remains fragmented, with limited insight into how privacy, resilience, and autonomous intelligence interact within integrated AISAC architectures. The motivation of this survey is to close that gap. We establish a forward-looking foundation for privacy-preserving and resilient AISAC systems, by systematically analyzing emerging threats, architectural challenges, enabling technologies, and future research directions across the converged domains of agentic AI and ISAC.
Addressing this gap requires more than cataloguing threats: it requires a framework for judging when a reviewed system is agentic, rather than merely AI-assisted, and for locating where in the perception--reasoning--action pipeline privacy, resilience, and other cross-cutting requirements apply. This survey makes the following contributions:
\begin{itemize}
\item We propose a \emph{six-stage closed-loop framework} comprising observation, contextualization, reasoning and prediction, planning and orchestration, execution and collaboration, and feedback and resilience. The framework integrates previously fragmented research on ISAC, RIS, UAVs, federated learning (FL), and agentic AI into a coherent technical narrative.
\item We introduce \emph{five levels of agentic maturity}, from physical-layer primitives to fully closed-loop agentic ISAC, and use them to classify representative studies by their primary stage, contribution side, and demonstrated (rather than claimed) agentic capability.
\item We conduct an \emph{audit of representative studies against nine agentic-specific evaluation criteria}, including contextual correctness, goal completion, tool-call success, and recovery time, and show that none of the audited studies reports more than one or two of them, exposing a gap between claimed and demonstrated agentic maturity.
\item Using this framework, we analyze privacy, security, resilience, and sustainability as cross-cutting requirements that apply differently at each stage of the closed loop, rather than as a single undifferentiated attack surface.
\end{itemize}

\begin{table*}[t]
\small 
\centering
\caption{Key Abbreviations Used in This Survey}
\label{tab:abbreviations}
\renewcommand{\arraystretch}{1.05}
\setlength{\tabcolsep}{4pt}
\begin{tabular}{ll|ll}
\hline
\textbf{Abbr.} & \textbf{Definition} &
\textbf{Abbr.} & \textbf{Definition} \\
\hline
6G      & Sixth-Generation Wireless Networks &
MARL    & Multi-Agent Reinforcement Learning \\

AI      & Artificial Intelligence &
MIMO    & Multiple-Input Multiple-Output \\

AISAC   & Agentic AI-enabled Integrated Sensing and Communication &
MPC     & Model Predictive Control \\

AoA     & Angle of Arrival &
MR      & Mixed Reality \\

AR      & Augmented Reality &
NOMA    & Non-Orthogonal Multiple Access \\

CIoT    & Consumer Internet of Things &
O-RAN   & Open Radio Access Network \\

CRB     & Cramér--Rao Bound &
RAN     & Radio Access Network \\

CSI     & Channel State Information &
RHS     & Reconfigurable Holographic Surface \\

DDPG    & Deep Deterministic Policy Gradient &
RIS     & Reconfigurable Intelligent Surface \\

DFRC    & Dual-Functional Radar and Communication &
SCA     & Successive Convex Approximation \\

DRL     & Deep Reinforcement Learning &
SLAM    & Simultaneous Localization And Mapping \\

FEEL    & Federated Edge Learning &
SNR     & Signal-to-Noise Ratio \\

FL      & Federated Learning &
STAR-RIS & Simultaneously Transmitting and Reflecting RIS \\

IoD     & Internet of Drones &
THz     & Terahertz \\

IoRT    & Internet of Robotic Things &
ToA     & Time of Arrival \\

IoT     & Internet of Things &
UAV     & Unmanned Aerial Vehicle \\

IoV     & Internet of Vehicles &
V2X     & Vehicle-to-Everything \\

ISAC    & Integrated Sensing and Communication &
VR      & Virtual Reality \\

LLM     & Large Language Model &
XR      & Extended Reality \\
\hline
\end{tabular}
\end{table*}

The remainder of this paper is organized as follows.
Section~\ref{sec:lr} reviews existing relevant surveys on ISAC, intelligent wireless systems, federated intelligence, and Agentic AI.
Section~\ref{sec:framework} introduces a unified technical framework for Agentic AI-enabled ISAC, 
and the proposed agentic maturity taxonomy.
Section~\ref{sec:physical_substrate} examines the physical observation and execution substrate.
Section~\ref{sec:contextual_intelligence} discusses how physical and multimodal measurements are transformed into contextual and semantic representations that support environmental understanding and task-aware decision-making.
Section~\ref{sec:reasoning_planning} reviews prediction, world modeling, goal interpretation, reasoning, and joint planning and orchestration across sensing, communication, and computation.
Section~\ref{sec:distributed_execution} investigates distributed and multi-agent execution.
Section~\ref{sec:trust_resilience} analyzes feedback, trust, privacy, security, and resilience across AISAC closed loop.
Section~\ref{sec:future_agentic} identifies open research challenges.
Section~\ref{sec:conclusion} concludes the paper. 
Table~\ref{tab:abbreviations} summarizes the main abbreviations used throughout the survey.

\section{Review of Relevant Surveys}
\label{sec:lr}
This section reviews the existing surveys on ISAC, AISAC, and federated privacy-preserving intelligence.
To position the contribution of this survey, Table~\ref{tab:survey_comparison} compares it with representative surveys from multiple perspectives.

\begin{table*}[t]
\centering
\caption{Comparison of representative surveys related to ISAC, intelligent wireless systems, and agentic AI.}
\label{tab:survey_comparison}
\resizebox{\textwidth}{!}{
\begin{tabular}{lcccccccc}
\toprule
\textbf{Survey} & \textbf{ISAC} & \textbf{AI/ML} &\textbf{ RIS/Edge AI} & \textbf{Agentic AI} & \textbf{Federated Intelligence} & \textbf{Privacy \& Security} & \textbf{Resilience} & \textbf{Open Challenges} \\
\midrule
Zhu et al. \cite{Zhu2025CST} & \checkmark & \checkmark & -- & -- & -- & \checkmark & Partial & \checkmark \\
Luong et al. \cite{Luong2025CST} & \checkmark & \checkmark & -- & -- & -- & Partial & -- & \checkmark \\
Ald et al. \cite{Ald2025Access} & \checkmark & \checkmark & RIS & -- & -- & \checkmark & -- & \checkmark \\
Zhang et al. \cite{zhang2025ArXiv} & \checkmark & Partial & -- & -- & -- & -- & -- & \checkmark \\
Kaushik et al. \cite{kaushik2023ArXiv} & \checkmark & -- & -- & -- & -- & Partial & -- & \checkmark \\
Chopra et al. \cite{chopra2025ris} & \checkmark & Partial & RIS & -- & -- & Partial & -- & \checkmark \\
Liu et al. \cite{Liu2025CST} & \checkmark & \checkmark & Edge AI & -- & Partial & Partial & -- & \checkmark \\
Wu et al. \cite{Wu2025IoTj} & \checkmark & \checkmark & Partial & Partial & \checkmark & Partial & Partial & \checkmark \\
Jiang et al. \cite{Jiang2025ArXivAI} & Partial & \checkmark & -- & \checkmark & -- & Partial & -- & \checkmark \\
Lu et al. \cite{lu2025agentic} & Partial & \checkmark & Edge AI & \checkmark & -- & Partial & -- & \checkmark \\
Acharya et al. \cite{Acharya2025Access} & -- & \checkmark & -- & \checkmark & -- & -- & -- & \checkmark \\
\midrule
\textbf{This Survey} &
\checkmark &
\checkmark &
\checkmark &
\checkmark &
\checkmark &
\checkmark &
\checkmark &
\checkmark \\
\bottomrule
\end{tabular}}
\end{table*}

\subsection{Applying ISAC to 6G}
\begin{figure}[t]
\begin{center}
    \includegraphics[width=0.6\columnwidth]{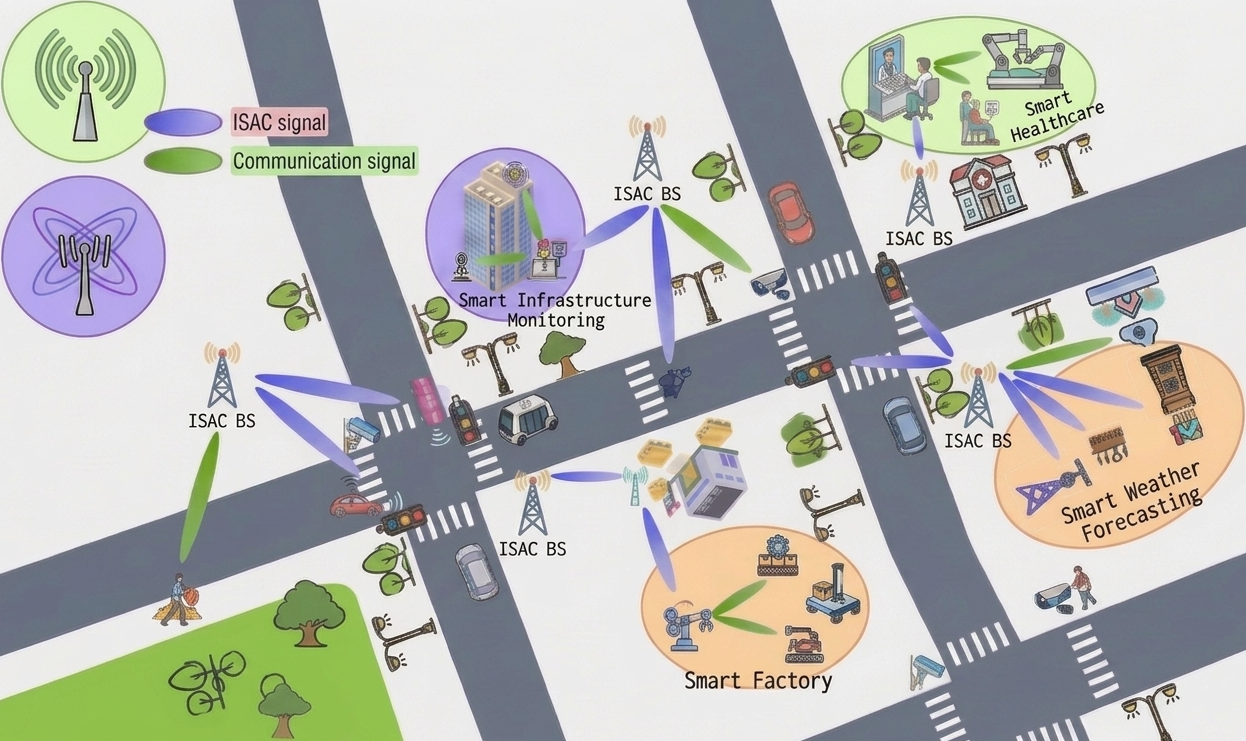}
\end{center}
\vspace{-1mm}
\caption{\small{Learning-driven ISAC frameworks can be applied to different tasks, such as joint waveform and beamforming design, environment-aware sensing, resource allocation, target detection and tracking, and cross-layer optimization~\cite{Luong2025CST}.}}
\vspace{-2mm}
\label{fig:0}
\end{figure}

In~\cite{Zhu2025CST}, ISAC architectures, enabling technologies, security vulnerabilities, and defenses were studied in emerging 6G. State-of-the-art cryptographic, physical-layer, signal-processing, and AI-driven protection techniques for ISAC systems were reviewed. \cite{Zhu2025CST} also identified open research problems, such as trustworthy joint waveform design, privacy-preserving sensing, secure beam management, and resilient ISAC-enabled autonomous systems. Other surveys have examined how learning can be embedded into ISAC design. According to~\cite{Luong2025CST}, learning-driven ISAC frameworks can be categorized by key tasks, including joint waveform and beamforming design, environment-aware sensing, resource allocation, target detection and tracking, and cross-layer optimization. As depicted in Fig.~\ref{fig:0}, \cite{Luong2025CST} examined model robustness, real-time constraints, and multi-agent coordination in learning-driven ISAC. It also analyzed research opportunities for applying ISAC intelligence to autonomous vehicles and UAV networks.

A parallel line of work has focused on the physical-layer technologies that make advanced ISAC possible. In~\cite{Ald2025Access}, ISAC in 6G was reviewed by examining the foundational principles enabling terahertz (THz) communications, RIS, and massive multiple-input multiple-output (MIMO). The authors of \cite{Ald2025Access} presented security and privacy challenges inherent to ISAC's dual-functional operation and outlined attack vectors across both domains. It also explained the growing role of AI in improving ISAC performance, covering learning-based beamforming, sensing parameter estimation, joint optimization, and autonomous decision-making. The authors in~\cite{Xu2023springer} provided an overview of ISAC in conjunction with THz communications, RIS, and massive MIMO. They analyzed the technical challenges associated with high-frequency propagation, hardware constraints, cross-layer coordination, and real-time signal processing.

\begin{figure*}[t]
\begin{center}
    \includegraphics[width=1.0\textwidth]{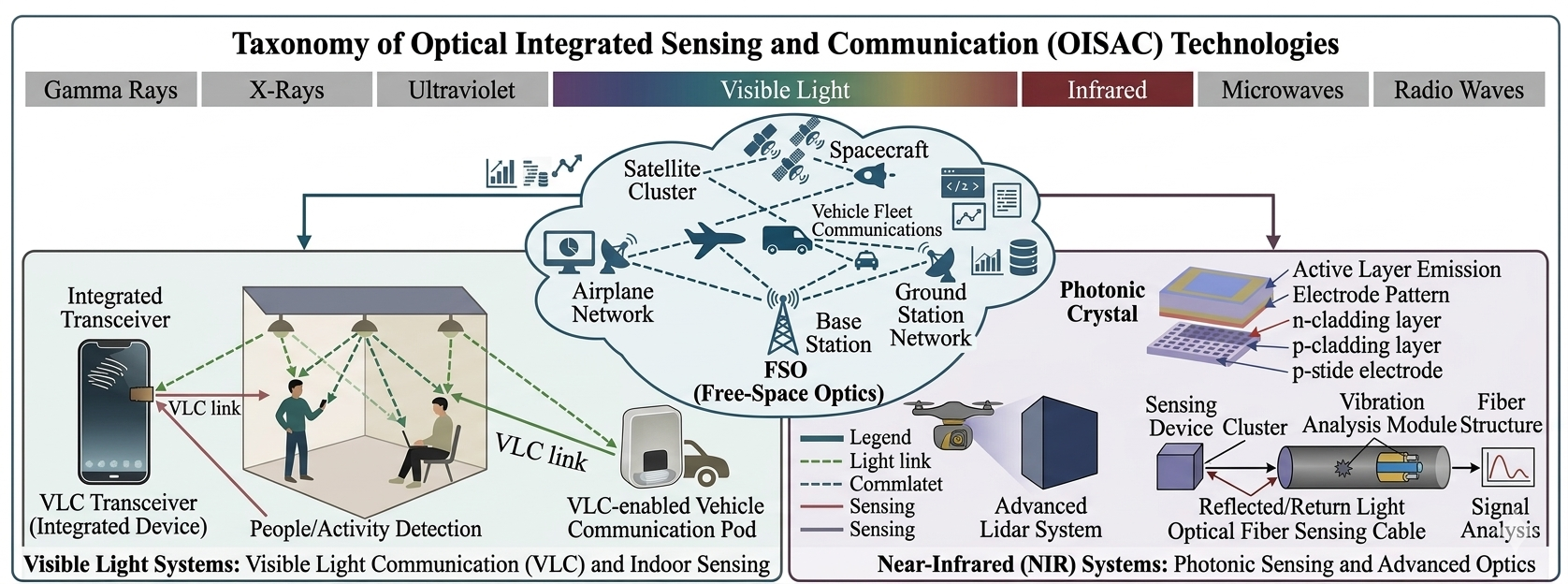}
\end{center}
\vspace{-1mm}
\caption{\small{Evolution of ISAC from radar-communication coexistence toward fully integrated AI-driven 6G architectures, where milestones include dual-functional waveforms, sensing-assisted communications, communication-assisted sensing, and unified ISAC~\cite{zhang2025ArXiv}.}}
\vspace{-2mm}
\label{fig:2}
\end{figure*}

Other surveys have taken a longer historical view. Zhang et al.~\cite{zhang2025ArXiv} provided an evolutionary perspective on ISAC, tracing its development from early radar-communication coexistence frameworks to today's deeply unified, AI-driven 6G ISAC architectures. As described in Fig.~\ref{fig:2},~\cite{zhang2025ArXiv} presented a timeline of key breakthroughs, including dual-functional waveforms, sensing-assisted communications, communication-assisted sensing, and fully integrated ISAC systems.

Standardization efforts complete this picture. Kaushik et al.~\cite{kaushik2023ArXiv} focused on the standardization of ISAC, where standardization organizations are establishing the technical, regulatory, and architectural foundations of ISAC for 6G. They surveyed ongoing and emerging standardization activities across major bodies, including 3GPP, ETSI, ITU-R, and IEEE, and discussed how ISAC is incorporated into sidelink-based sensing, joint waveform design, and massive MIMO. \cite{kaushik2023ArXiv} identified technical requirements, use cases, and performance metrics needed to address legacy systems coexistence, spectrum allocation, and privacy considerations.

\subsection{Intelligent and Reconfigurable ISAC: From RIS to Edge-AI Integration}
RIS-assisted ISAC has emerged as a promising paradigm for 6G, enabling networks to shape the radio environment to enhance sensing accuracy and communication efficiency.
An examination of RIS-assisted ISAC was conducted in~\cite{chopra2025ris}, focusing on how RIS can enhance dual-functional wireless systems by manipulating the radio environment. It was showed that RIS-enabled ISAC can improve sensing resolution and expand coverage for autonomous mobility, smart city monitoring, etc. Several challenges arise, including joint optimization of RIS configurations, hardware constraints, real-time control, interference management, and robust signal processing under dynamic conditions.

Related work has examined RIS specifically for localization. The authors in~\cite{Umer2025CST} surveyed the application of RIS to 6G radio localization, and reviewed RIS-assisted localization techniques, including geometry-aware channel modeling, beam manipulation, angle and delay estimation, and cooperative sensing.
Meng et al.~\cite{Meng2025Network} extended this direction by exploring the convergence of ISAC with smart propagation engineering, where RIS and holographic MIMO jointly shape wireless environments to enhance sensing accuracy and communication efficiency. Since RIS-assisted ISAC can support environment-aware beamforming, high-resolution sensing, joint localization and communication, and adaptive coverage optimization, practical challenges, such as real-time control, hardware imperfections, channel modeling complexity, and integration into existing network architectures, warrant further study.

In~\cite{Liu2025CST}, ISAC architectures that incorporate distributed edge intelligence were reviewed. The fusion of ISAC with edge AI is essential for achieving real-time, intelligent perception in 6G; it enables distributed, low-latency, and privacy-preserving analysis at the network edge. 
This edge-intelligence perspective extends to consumer applications. ISAC-driven multimodal and heterogeneous data integration was developed for consumer Internet of Things (IoT)~\cite{Chen2025TCE}, where sensing and communication are jointly designed to support multiple data types, such as audio, vision, environmental, and device-state information. 

Deep learning has been studied as a unifying tool across these efforts. Temiz et al.~\cite{Temiz2025JCS} surveyed deep learning for ISAC systems, where neural networks model joint waveform design, target detection, channel estimation, interference mitigation, and cross-domain optimization. Supervised, unsupervised, and reinforcement learning approaches were categorized to examine complex sensing-communication features. While deep learning can unlock advanced joint optimization capabilities for ISAC, it still requires addressing limitations in robustness, generalization, interpretability, and real-time deployment across heterogeneous wireless environments.

\subsection{Emergence of Agentic AI}
\begin{figure}[t]
\begin{center}
    \includegraphics[width=0.60\columnwidth]{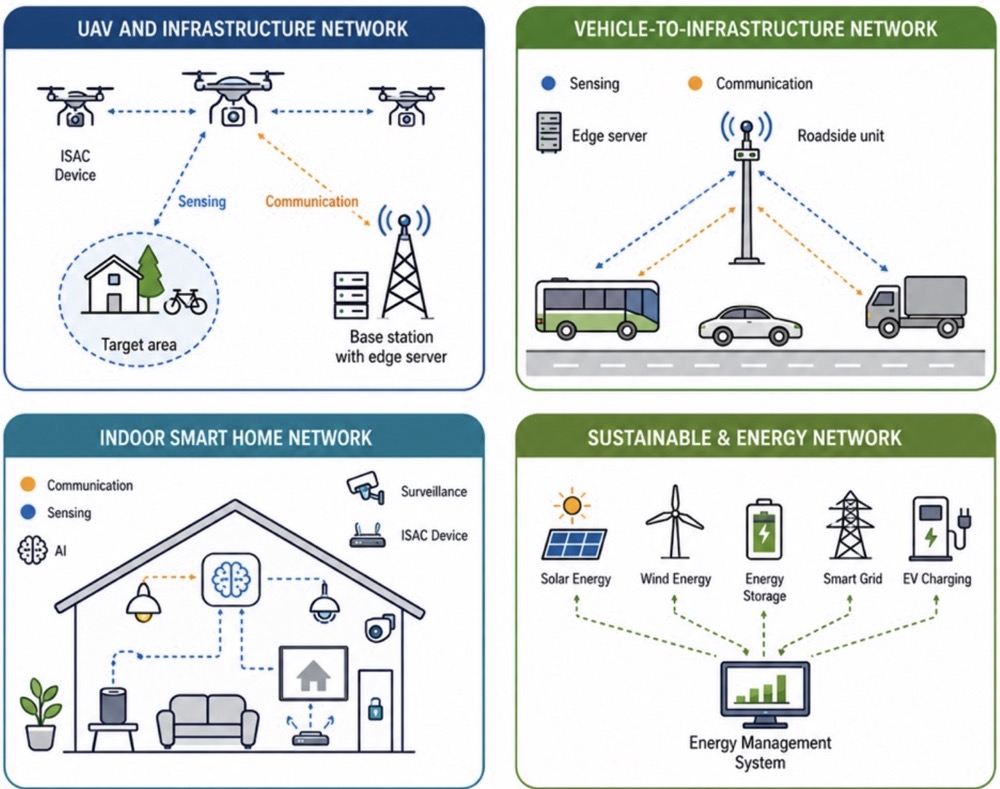}
\end{center}
\vspace{-1mm}
\caption{\small{AISAC as a closed-loop intelligence fabric, leverages agentic AI techniques, including deep learning for sensing interpretation, reinforcement learning for resource scheduling, federated and distributed learning for privacy preservation, and multi-agent learning for cooperative perception and control~\cite{Wu2025IoTj}.}}
\vspace{-2mm}
\label{fig:3}
\end{figure}

Fig.~\ref{fig:3} illustrates that agentic AI has emerged as a new paradigm, shifting the focus from passive, model-centric intelligence to autonomous systems capable of reasoning, planning, and acting to achieve complex goals.
Intelligent communications can transition from model-centric intelligence based on large AI models to agent-centric intelligence~\cite{Jiang2025ArXivAI}, where autonomous, event-driven agents orchestrate learning and control across the network. Foundation models, large language models (LLMs), and multimodal AI were examined as components that can be embedded into communication systems to enable semantic-aware transmission, autonomous network management, cross-layer optimization, and human-machine interaction. It was noted in \cite{Jiang2025ArXivAI} that agentic AI in ISAC requires rethinking network architectures and protocol design to support reasoning, collaboration, trust, and continual learning at scale.

The authors in~\cite{Acharya2025Access} focused on agentic AI-enabled autonomous intelligent systems, which support long-horizon planning, self-directed goal decomposition, and closed-loop interaction with complex environments. \cite{Acharya2025Access} reviewed algorithmic foundations spanning hierarchical reinforcement learning, imitation learning, LLM-based reasoning and planning, (e.g., chain-of-thought prompting and tool use), and hybrid symbolic-neural approaches.
These ideas have already begun to merge with ISAC. In~\cite{Wu2025IoTj}, AI-enabled ISAC was presented, framing ISAC as a closed-loop intelligence fabric rather than a loose coupling of individual functions. \cite{Wu2025IoTj} classified AI applications in ISAC, including deep learning for sensing signal interpretation, reinforcement learning for resource scheduling, federated and distributed learning for privacy protection, and multi-agent learning for cooperative perception and control.

Agentic AI has also been positioned more broadly as a key enabler for edge general intelligence in future 6G systems~\cite{lu2025agentic}. In \cite{lu2025agentic}, agentic AI architectures were categorized between single-agent and multi-agent settings, detailing how perception, reasoning, and resource control can be integrated to support tasks such as spectrum management, power control, routing, edge inference offloading, and cooperative sensing in ISAC.
In~\cite{alshalwi2025toward}, agentic AI was studied with bandwidth prioritization for distributed 6G ISAC systems, using autonomous decision-making to manage competing sensing and communication demands. Bandwidth allocation was formulated as a distributed, goal-driven control problem, in which intelligent agents observe local network states, reason about sensing-communication trade-offs, and prioritize bandwidth to meet heterogeneous latency, reliability, and perception requirements.

\subsection{Positioning of This Survey}

\textit{From functional integration to intelligent, federated ISAC.} Existing ISAC surveys for 6G primarily focus on the functional co-design of sensing and communication, and more recently on their integration with computation, AI/ML, RIS, and edge intelligence. A key lesson is that future ISAC systems, especially in consumer IoT (CIoT), UAV swarms, and vehicular networks, cannot rely on monolithic intelligence, due to scalability, latency, and data ownership constraints. This motivates our survey to position federated intelligence as a new paradigm for ISAC, one that enables distributed learning and decision-making across sensing and communication domains without centralized data aggregation.

\textit{Insufficient security awareness without privacy- and resilience-by-design.} The surveyed literature increasingly acknowledges security challenges in ISAC, including sensing spoofing, waveform manipulation, RIS attacks, and AI vulnerabilities. Privacy preservation, robustness against adversarial behaviors, and resilience under partial failures or Byzantine agents are often treated as isolated problems. Our survey aims to reveal a critical gap: ISAC systems that tightly couple sensing, communication, and AI amplify attack surfaces and privacy leakage risks, especially when multimodal data and edge intelligence are involved.

\textit{Agentic, resilient federated intelligence can be developed for ISAC.} Recent surveys on agentic AI describe a paradigm shift toward autonomous, goal-driven intelligence with reasoning, planning, and coordination capabilities, yet their integration with ISAC remains unexplored. Meanwhile, ISAC surveys lack a unifying intelligence model capable of long-horizon decision-making, self-adaptation, and cooperative control under uncertainty. The lesson is clear: next-generation ISAC requires agentic, federated intelligence that operates across distributed nodes, learns collaboratively under privacy constraints, and remains resilient to dynamic environments and adversarial conditions.


\section{A Unified Technical Framework for Agentic AI-Enabled ISAC}
\label{sec:framework}

Existing research has separately investigated ISAC, learning-driven resource orchestration, multimodal environmental understanding, distributed intelligence, and autonomous network control. Joint beamforming and computation-offloading frameworks establish the physical substrate for tightly coupled ISAC operation~\cite{Zhao2025TC,Chen2024Sensor,Liu2024TWC}, while federated and privacy-aware designs extend this substrate toward collaborative intelligence~\cite{Hu2025TWC,Li2025CM,Yan2023IoTj}. In parallel, reinforcement-learning-based control, intent-driven network automation, and LLM-enabled multi-agent architectures have introduced adaptive control, workflow composition, and autonomous reasoning into wireless networks~\cite{Pala2025TWC,Wu2024ICC,LI2025Engineering,coelho2025a4fn,luo2025ai}. 
Synthesizing these research directions, this survey organizes the literature according to the following technical progression: physical observation $\rightarrow$ contextualization $\rightarrow$ prediction and reasoning $\rightarrow$ planning and orchestration $\rightarrow$ execution and collaboration $\rightarrow$ feedback and resilience.
This progression provides a framework for identifying what each study contributes on the ISAC side, what it contributes on the AI side, and which capabilities remain necessary for full agentic operation.

\subsection{From Physical-Layer ISAC to Closed-Loop Agentic Intelligence}

ISAC provides a shared physical-layer infrastructure for environmental sensing and wireless connectivity~\cite{Zhao2025TC,Chen2024Sensor}. Several related studies extend this foundation by explicitly incorporating edge computation, task offloading, and over-the-air computation~\cite{Liu2024TWC,Li2023TWC}. In this survey, these computation-aware mechanisms are included when they enable the ``perception-reasoning-action'' loop of AISAC. Existing ISAC and computation-aware studies jointly optimize sensing accuracy, communication rate, transmit power, and computation workloads,  exposing a set of controllable physical and computational resources that can later be coordinated by an autonomous agent~\cite{Zhao2025TC,Liu2024TWC}.

Physical measurements alone do not fully reveal the operational meaning of the sensed environment. For example, a moving reflector can be characterized by range and Doppler, but these measurements do not directly indicate whether the object is a vehicle, whether its behavior is anomalous, or whether it should receive additional sensing and communication resources. Learning-enabled ISAC, multimodal sensing, holographic localization, and digital-twin-assisted prediction begin to bridge this gap by transforming signal-level measurements into task-relevant environmental representations~\cite{Hu2023JSAC,Li2025CM,coelho2025a4fn}.

Agentic AI extends this contextual intelligence into goal-directed closed-loop operation. Recent agent-oriented architectures allow network agents to interpret intent, compose workflows, coordinate specialized agents, and generate network actions~\cite{LI2025Engineering,coelho2025a4fn,zhang2025toward}. Related studies on agentic graph neural networks, foundation-model agents, and reasoning for wireless systems demonstrate the role of structured reasoning, knowledge transfer, and cross-layer decision-making~\cite{lu2025agentic,xiao2025towards,luo2025ai}. Based on these developments, the closed loop can be abstracted as
\begin{equation}
 y_t^{\mathrm{RF}}
 \rightarrow o_t^{\mathrm{phy}}
 \rightarrow c_t^{\mathrm{sem}}
 \rightarrow \hat{s}_{t+1:t+H}
 \rightarrow p_t
 \rightarrow a_t
 \rightarrow y_{t+1}^{\mathrm{RF}},
\label{eq:agentic_loop}
\end{equation}
where $y_t^{\mathrm{RF}}$ denotes received communication and sensing signals, $o_t^{\mathrm{phy}}$ denotes extracted physical observations, $c_t^{\mathrm{sem}}$ denotes contextual and semantic state, $\hat{s}_{t+1:t+H}$ denotes predicted future states over horizon $H$, $p_t$ denotes an agent-generated plan, and $a_t$ denotes sensing, communication, computation, mobility, and infrastructure-control actions.

Equation~\eqref{eq:agentic_loop} is a synthesis introduced in this survey rather than a model adopted from a single prior work. Its individual stages are supported by different strands of literature: physical ISAC and computation-aware optimization~\cite{Zhao2025TC,Liu2024TWC}, multimodal contextualization~\cite{Hu2023JSAC,coelho2025a4fn}, adaptive and multi-agent control~\cite{Pala2025TWC,Wu2024ICC}, intent-driven orchestration~\cite{LI2025Engineering,luo2025ai}, and resilient distributed learning~\cite{Yan2023IoTj,Zhang2025TITS}.

\subsection{Functional Roles of ISAC, AI, and Autonomous Agents}

The convergence of Agentic AI and ISAC involves three related but distinct functional layers, with computation acting as a supporting capability across them. Distinguishing them is important because the use of AI, FL, Deep Reinforcement Learning (DRL), or Multi-Agent Reinforcement Learning (MARL) does not by itself imply that a system is agentic.

\subsubsection{Physical ISAC and Supporting Computation Layer}

The physical ISAC and supporting computation layer performs signal transmission, sensing, communication, and low-level control, while edge and distributed computation support inference, task execution, and resource coordination. Representative functions include waveform design, target detection and localization, beamforming, power and spectrum control, RIS configuration, over-the-air aggregation, computation offloading, and edge execution~\cite{Zhao2025TC,Chen2024Sensor,Liu2024TWC,Kesargheh2025TVT}. Its observations may be represented as
\begin{equation}
\begin{aligned}
 o_t^{\mathrm{phy}}=
 \{
 &\mathrm{CSI},\mathrm{AoA},\mathrm{ToA},\mathrm{Doppler},
 \mathrm{range},\mathrm{velocity}, 
 \mathrm{SNR},\mathrm{confidence}\}.
\end{aligned}
\label{eq:physical_obs}
\end{equation}
where CSI denotes channel state information, AoA and ToA denote angle of arrival and time of arrival, respectively, and SNR denotes signal-to-noise ratio.
Most optimization-centric ISAC and computation-aware studies primarily contribute to this layer. Their beamforming, scheduling, and offloading algorithms are therefore best interpreted as executable tools for a higher-level agent rather than as complete agents themselves.

\subsubsection{AI-Based Contextual Intelligence Layer}

The contextual intelligence layer converts physical observations into task-relevant representations. Multimodal architectures can combine RF sensing with camera, LiDAR, GPS, topology, mobility, and historical information, while federated and transfer-learning mechanisms can integrate knowledge across heterogeneous devices without centralizing raw data~\cite{Hu2023JSAC,Hu2025TWC,coelho2025a4fn}. The resulting context may be represented as
\begin{equation}
\begin{aligned}
 c_t^{\mathrm{sem}}=
 \{
 &\text{object identity},\text{activity},\text{risk},
 \text{scene state},
\text{network state},\text{uncertainty}\}.
\end{aligned}
\label{eq:semantic_state}
\end{equation}
This layer can include multimodal fusion, semantic mapping, anomaly detection, uncertainty estimation, task-relevance assessment, and world-model construction. The A4FN architecture~\cite{coelho2025a4fn} separates multimodal perception from decision and action, illustrating how environmental observations can be contextualized before network reconfiguration.

\subsubsection{Agentic Reasoning and Orchestration Layer}

The agentic layer operates on context together with mission goals, memory, predicted states, and available resources. A generic agent state can be expressed as
\begin{equation}
 s_t^{\mathrm{agent}}=
 \{c_t^{\mathrm{sem}},g_t,m_t,\hat{s}_{t+1:t+H},\mathcal{R}_t\},
\label{eq:agent_state}
\end{equation}
where $g_t$ is the current objective, $m_t$ is memory, and $\mathcal{R}_t$ denotes sensing, communication, computation, energy, and mobility resources. Agent-oriented network architectures have begun to support intent interpretation, workflow composition, reasoning, and coordinated action~\cite{LI2025Engineering,coelho2025a4fn,luo2025ai}. The resulting action may be written as
\begin{equation}
 a_t=\{a_t^{\mathrm{sense}},a_t^{\mathrm{comm}},a_t^{\mathrm{comp}},
 a_t^{\mathrm{mobility}},a_t^{\mathrm{infra}}\},
\label{eq:agent_action}
\end{equation}
including sensing schedules, spectrum and beam configurations, computation-offloading decisions, UAV or vehicle trajectories, and RIS configurations.

\subsection{Why Agentic AI Rather Than Conventional AI-Based ISAC?}

Conventional AI has been applied to sensing interpretation, channel estimation, beamforming, resource allocation, FL, and adaptive control~\cite{Hu2025TWC,Pala2024GLOBECOM,Pala2025TWC,Wu2024ICC}. Agentic AI should not be justified by claiming that conventional AI is necessarily static.
The distinction instead lies in the scope, representation, and organization of the decision process.

\begin{table*}[t]
\centering
\caption{Conventional AI-enabled ISAC versus Agentic AI-enabled ISAC.}
\label{tab:conventional_agentic}
\renewcommand{\arraystretch}{1.12}
\resizebox{\textwidth}{!}{%
\begin{tabular}{p{2.1cm}p{6cm}p{7.5cm}}
\toprule
\textbf{Dimension} & \textbf{Conventional AI-enabled ISAC} & \textbf{Agentic AI-enabled ISAC} \\
\midrule
Objective & Predefined loss, reward, or optimization target & High-level mission or operator intent that must be translated into technical objectives \\
Input & Task-specific observations or fixed state vectors & Physical observations, semantic context, goals, memory, predictions, and resource states \\
Decision & Prediction, classification, or one-step control action & Multi-step plan coordinating multiple functions, tools, and entities \\
Scope & A single function such as beamforming, detection, or offloading & Cross-layer sensing, communication, computation, mobility, and infrastructure orchestration \\
Execution & Direct model output or a dedicated optimizer & Selection and invocation of models, solvers, controllers, digital twins, and network tools \\
Collaboration & Fixed protocol, FL round, or learned joint policy & Dynamic role assignment, capability discovery, negotiation, and team formation \\
Adaptation & Model retraining or policy update & Outcome evaluation, reflection, tool switching, replanning \\
Failure handling & Predefined robust objective or fallback policy & Diagnosis, recovery planning, task migration, isolation, escalation \\
\bottomrule
\end{tabular}}
\end{table*}

For example, an optimization-based ISAC controller may maximize a weighted communication--sensing utility under predefined constraints, as in joint beamforming and resource-allocation studies~\cite{Zhao2025TC,Liu2024TWC}. By contrast, an agentic system may receive the mission-level objective of maintaining reliable localization for safety-critical vehicles while preserving minimum connectivity for other users. It must identify relevant entities, translate the intent into sensing and communication requirements, predict blockage and mobility, select cooperating nodes, invoke suitable beamforming or scheduling tools, and revise its plan when the outcome is unsatisfactory.
In Table~\ref{tab:conventional_agentic}, we compare the conventional and agentic AI-enabled ISAC from different dimensions.

\subsection{Hierarchical and Multi-Timescale Agent--ISAC Architecture}
\label{sec:framework:hier}

A practical architecture should not use a computationally intensive reasoning model to control every symbol or transmission slot. Existing ISAC optimizers and learned controllers are often designed for fast resource adaptation~\cite{Pala2025TWC,Wu2024ICC}, whereas intent interpretation and workflow composition operate at slower task or mission time scales~\cite{LI2025Engineering,luo2025ai}. This motivates a hierarchy with three interacting loops.
\begin{itemize}
    \item 
\textit{Fast physical-control loop:} Deterministic signal-processing algorithms, optimizers, model predictive control (MPC) controllers, or lightweight policies perform beam tracking, power control, waveform adaptation, RIS updates, packet scheduling, and safety-critical mobility control.
    \item 
\textit{Medium-timescale orchestration loop:} Specialized agents determine sensing priorities, target schedules, cooperating nodes, computation offloading, resource budgets, and collaboration topology. Multi-agent F-DDPG (Deep Deterministic Policy Gradient) and federated Q-learning exemplify adaptive mechanisms that can support this level~\cite{Pala2025TWC,Wu2024ICC}.
    \item 
\textit{Slow cognitive and learning loop:} Mission agents perform intent interpretation, long-horizon planning, world-model updating, continual learning, memory consolidation, and system-level adaptation~\cite{LI2025Engineering,coelho2025a4fn,zhang2025toward}.    
\end{itemize}
The hierarchy can be summarized as Mission Agent $\rightarrow$ ISAC Orchestrator $\rightarrow$ Schedulers and Solvers $\rightarrow$ Physical Controllers.
This separation allows agentic reasoning to coordinate real-time ISAC mechanisms without replacing them.

\subsection{Agent Roles and Tool Interfaces}
\label{sec:framework:agent}

Future systems will rely on specialized rather than monolithic intelligence. Agentic network cores and flying-network architectures already distinguish intent, perception, reasoning, and action functions~\cite{LI2025Engineering,coelho2025a4fn}. Based on these developments, representative roles include a mission agent for intent translation, a context agent for multimodal fusion, a prediction agent for mobility and channel forecasting, sensing and communication agents for resource scheduling, a computation agent for edge execution, an infrastructure agent for RIS and UAV control, and a security agent for verification and recovery.
These agents need not directly solve every low-level problem. Instead, they can invoke waveform and beamforming solvers, RIS optimizers, spectrum schedulers, channel predictors, digital twins, DRL policies, offloading engines, secure aggregation mechanisms, and anomaly detectors. This tool-oriented interpretation allows optimization-centric ISAC and computation-aware studies~\cite{Zhao2025TC,Liu2024TWC,Kesargheh2025TVT}, adaptive controllers~\cite{Pala2025TWC,Wu2024ICC}, and intent-driven agent architectures~\cite{LI2025Engineering,luo2025ai} to be placed in one coherent framework.

\subsection{Literature Alignment and Agentic Maturity}
\label{sec:framework:literature_alignment}
To avoid treating every use of AI, FL, DRL, or MARL as Agentic AI, we classify each study along three dimensions: its primary stage in the closed loop, its main contribution side, and its agentic maturity.

\subsubsection{Technical-Story Stage}
Each study is assigned one primary stage: \textbf{O} (observation), \textbf{C} (contextualization), \textbf{R} (reasoning and prediction), \textbf{P} (planning and orchestration), \textbf{E} (execution and collaboration), or \textbf{F} (feedback and resilience). A study may support multiple stages, but assigning a primary role prevents the same work from being repeatedly presented as a complete solution.

\subsubsection{Contribution Side}
A study is classified as \emph{ISAC-centric} when it mainly improves sensing, communication, computation, or physical control; \emph{AI-centric} when it mainly improves contextualization, prediction, reasoning, or learning; \emph{interface-centric} when it explicitly connects AI decisions with ISAC observations or actions; and \emph{end-to-end agentic} when it implements a substantial portion of the perception-reasoning-action-feedback loop.

\subsubsection{Agentic Maturity}
We define five maturity levels grounded in representative literature.

\textit{Level 0---Physical ISAC primitive:} waveform, beamforming, sensing, RIS, or computation mechanisms without an adaptive AI controller, represented by optimization-centric ISAC and computation-aware designs~\cite{Zhao2025TC,Chen2024Sensor}.

\textit{Level 1---AI-assisted ISAC component:} AI improves one task, such as sensing interpretation, model aggregation, resource allocation, or privacy-aware learning~\cite{Hu2025TWC,Liu2024TWC}.

\textit{Level 2---Adaptive ISAC controller:} the system observes network states and dynamically chooses actions through RL, MARL, or online learning~\cite{Pala2024GLOBECOM,Pala2025TWC,Wu2024ICC}.

\textit{Level 3---Context-aware agentic component:} the system introduces semantic context, intent interpretation, structured reasoning, tool selection, or explicit planning~\cite{LI2025Engineering,coelho2025a4fn,luo2025ai}.

\textit{Level 4---Closed-loop Agentic ISAC:} the system integrates physical or contextual perception, mission-level goals, memory and prediction, multi-step planning, tool-enabled execution, outcome evaluation, and feedback-driven adaptation. Existing agentic network and multi-agent frameworks approach parts of this level, but the reviewed literature does not yet provide a mature, widely validated end-to-end implementation spanning all stages~\cite{LI2025Engineering,coelho2025a4fn,zhang2025toward}.

\begin{table*}[t]
\centering
\caption{Representative literature across the Agentic AI--ISAC technical story.}
\label{tab:taxonomy}
\renewcommand{\arraystretch}{1.12}
\resizebox{\textwidth}{!}{%
\begin{tabular}{p{2.2cm}p{2.4cm}p{4.4cm}p{5.4cm}p{1.5cm}}
\toprule
\textbf{Rep. works} & \textbf{Primary stage} & \textbf{ISAC contribution} & \textbf{AI/agentic contribution} & \textbf{Level} \\
\midrule
\cite{Zhao2025TC,Chen2024Sensor} & O/E & Joint sensing, communication, computation, and physical resource control & No explicit semantic reasoning or goal-directed planning & L0 \\
\cite{Liu2024TWC,Hu2025TWC} & O/P/F & Learning-driven and privacy-aware ISAC & Task-specific learning and adaptive coordination & L1--L2 \\
\cite{Hu2023JSAC,coelho2025a4fn} & C & RF and multimodal environmental observations & Context construction, semantic perception, and adaptive interpretation & L2--L3 \\
\cite{Pala2025TWC,Wu2024ICC} & P/E & Power, beamforming, RIS, and multi-node action spaces & Federated or multi-agent adaptive control policies & L2 \\
\cite{LI2025Engineering,luo2025ai} & R/P/E & Network state and executable control interfaces & Intent interpretation, reasoning, workflow composition, and tool-oriented control & L3 \\
\cite{Yan2023IoTj,Zhang2025TITS} & F & Distributed learning and network operation under faults or attacks & Trust, privacy protection, and resilient adaptation & L1--L2 \\
\bottomrule
\end{tabular}}
\end{table*}

Table~\ref{tab:taxonomy} maps a set of representative studies onto these primary stages, contribution sides, and maturity levels, illustrating that most reviewed works cluster at L0--L2 and that L3--L4 systems remain comparatively rare.
Based on this framework, the remainder of the survey follows the progression of the closed loop. We first examine how physical ISAC provides perception and execution capabilities, how supporting computation enables timely inference and control, and how AI transforms physical measurements into contextual and semantic intelligence. We then study prediction, reasoning, goal decomposition, and joint sensing--communication--computation orchestration. Distributed execution is reviewed through FL, MARL, RIS, UAV, vehicular, and robotic systems, followed by feedback, continual adaptation, privacy, security, resilience, and sustainable operation. Every study is positioned according to the function it contributes while clarifying which capabilities are missing for complete closed-loop Agentic ISAC.


\section{Physical ISAC as the Observation and Execution Substrate}
\label{sec:physical_substrate}

This section positions the physical sensing, communication, and supporting computation mechanisms as the observation and execution substrate of the Agentic AI--ISAC loop. Its exposes measurable physical states and controllable resources, while higher-level contextualization, goal reasoning, and multi-step planning are generally outside their scope. Their optimization and control mechanisms can therefore be interpreted as tools that an agent may configure or invoke.

\subsection{Integrated Physical Observation and Task-Oriented Interfaces}

Recent advances in AISAC have improved sensing accuracy, communication efficiency, and resource utilization. Most existing designs, however, still rely on static or task-specific AI models that act as optimization or inference modules rather than autonomous decision-makers \cite{Cheng2026tnse}. Meanwhile, edge AI and task-oriented AISAC paradigms have revealed that sensing and communication are inherently decision-coupled processes. Their joint optimization cannot be addressed by fixed control policies or single-shot learning. This gap motivates the introduction of agentic AI, which endows AISAC with goal-driven reasoning, continual learning, and closed-loop interaction with the environment. 

Prior studies on AISAC, e.g., \cite{Wu2024CM}, demonstrate that embedding learning mechanisms into sensing and communication improves adaptability, spectral efficiency, and situational awareness. The AISAC framework for edge intelligence in~\cite{zhu2023springer} reveals that sensing and communication are intrinsically interdependent processes, whose joint design is guided by downstream task objectives. Over-the-air FL frameworks show that communication, computation, and learning performance are tightly coupled, and that communication-centric designs can yield suboptimal learning outcomes due to straggler effects, channel heterogeneity, and resource imbalance \cite{Liu2021TWC}. More recently, FL-based AISAC paradigms have extended this coupling by treating radio sensing as the data acquisition mechanism for distributed learning. They demonstrate that sensing quality, device participation, resource allocation, and privacy constraints jointly determine learning convergence and system performance~\cite{Hu2025TWC}. 

These works indicate that future AISAC systems increasingly operate as decision-intensive systems, in which sensing, communication, computation, and learning can continuously adapt to evolving environments and task requirements. Under this paradigm, sensing is elevated from passive observation into an intentional action for uncertainty reduction. Communication becomes a selective process for task-relevant information exchange. Computation supports continual planning and learning at the network edge. This principle-level integration transforms ISAC from an AI-assisted signal-processing platform into a self-directed, task-aware intelligent system, aligning with task-oriented AISAC and federated edge intelligence visions for scalable, resilient, and privacy-aware 6G networks.


\subsection{Resource Coordination and Executable ISAC Tools}

Once agentic decision-making becomes the organizing principle, the next question is how to coordinate resources at scale. As AISAC scales toward dense, heterogeneous, and edge-native deployments, the intelligent co-optimization of sensing and communication resources becomes a challenge. Conventional ISAC designs emphasize static beamforming or link-level trade-offs. Large-scale AISAC, by contrast, operates under multi-agent cooperation, non-IID data distributions, dynamic channels, and stringent energy and latency constraints, while preserving data privacy at the edge. 

\begin{figure}[t]
\begin{center}
    \includegraphics[width=0.60\columnwidth]{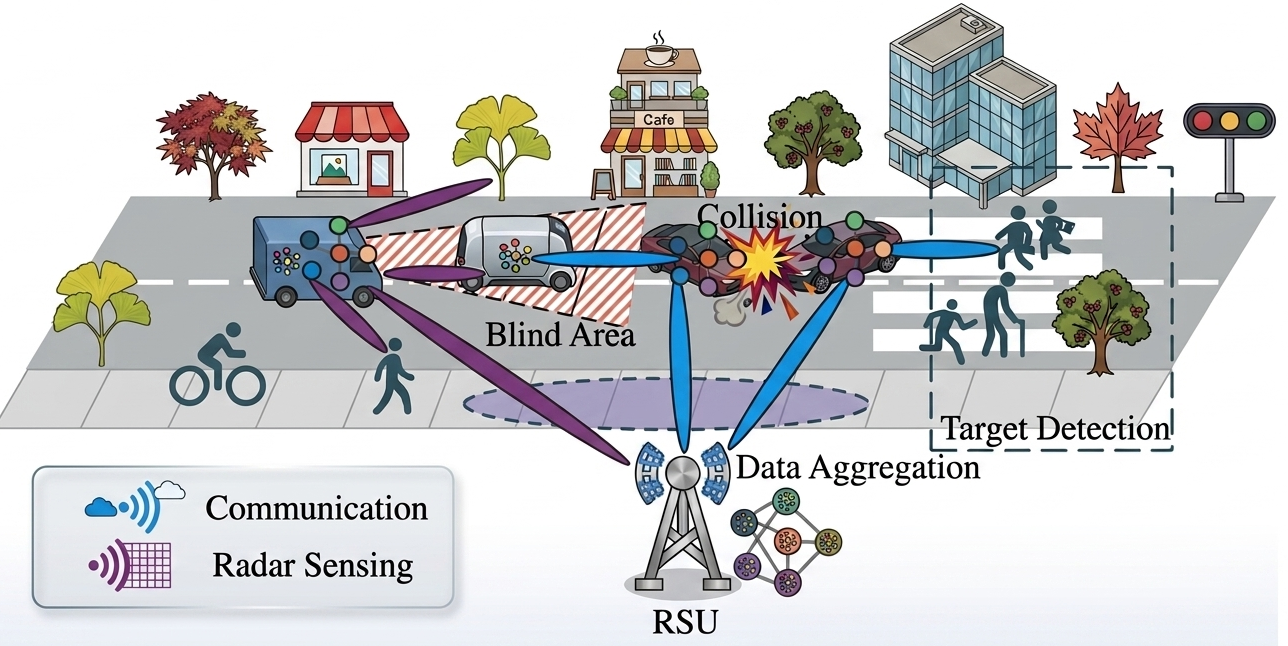}
\end{center}
\vspace{-1mm}

\caption{\small{Application of AISAC in vehicular network \cite{Yang2024TWC}.}}

\vspace{-2mm}
\label{fig:8}
\end{figure}

Ouyang et al. \cite{Ouyang2024SJ} present a communication-efficient FL framework that couples FL with reinforcement learning. It handles non-IID data without requiring clients to reveal global data distributions, by learning the optimal data augmentation under an explicit accuracy-overhead objective. The resulting design reduces communication burden through lightweight model construction. 
In vehicular networks, Yang et al. \cite{Yang2024TWC} formulate a multi-user MIMO AISAC system, illustrated in Fig.~\ref{fig:8}, where beamforming and power allocation balance the trade-offs among sensing and communication. They adopt DDPG for adaptive joint resource allocation, and study an AirComp-based FL scheme to support low-latency cooperation. Complementarily, task-oriented AISAC design for edge inference emphasizes that sensing, computation (feature extraction and quantization), and communication jointly determine downstream inference accuracy under latency constraints \cite{Wen2024TWC}. This motivates systematic joint allocation. AISAC resource coordination is therefore inherently task-driven, where the relevant performance metrics shift from throughput-centric objectives to latency, accuracy, and distortion trade-offs that depend on the entire sensing, compute, and transmit pipeline \cite{Wen2024TWC}.

Energy efficiency constitutes another defining dimension of edge-native AISAC, especially for resource-constrained devices. Kesargheh et al. \cite{Kesargheh2025TVT} develop an energy-efficient AISAC framework for privacy-preserving wireless FL. It optimizes user selection (guided by local computing time), RIS-assisted channel enhancement, and joint allocation of power, computing frequency, bandwidth, and sensing constraints, via a mixed-integer nonlinear formulation and iterative successive linear programming. At the physical layer, Li et al. \cite{Li2023TWC} describe the ISCCO framework, which unifies sensing and communication within a single transmission by combining AISAC with AirComp. This reveals intrinsic trade-offs between radar estimation accuracy and function-computation error, and requires joint beamforming designs to manage the coupling. These studies show that energy-aware AISAC should be built on joint structural choices, e.g., RIS assistance, multiple access, and over-the-air aggregation, aligned with sensing and computation objectives~\cite{Kesargheh2025TVT, Li2023TWC}.

At the scheduling and access level, adaptive coordination introduces additional degrees of freedom, which help avoid performance collapse in dense multi-target and multi-user regimes. Dou et al. \cite{Dou2023TVT} show that sensing and communication can both degrade when many targets are sensed simultaneously, motivating sensing scheduling in non-orthogonal multiple access (NOMA)-aided AISAC. By optimizing beamforming, NOMA transmission duration, and sensing-target scheduling under mutual-information-based sensing constraints, their framework improves sensing efficiency while maintaining communication guarantees.

For highly dynamic edge environments, e.g., vehicle-to-everything (V2X), Shang et al. \cite{shang2025ArXiv} model AISAC-enabled beamforming as a Markov decision process. This enables model-free adaptation without frequent CSI acquisition. Spiking neural networks are integrated into DRL to reduce energy consumption through sparse, event-driven computation. The result points to a practical pathway for green, learning-based AISAC control under continuous operation.


\subsection{Programmable Radio Environments and RIS}
\label{sec:physical_substrate:radio}


Several recent surveys provide a foundation for understanding RIS technologies and their evolution. Iqbal et al.~\cite{iqbal2025comprehensive} review the principles of RIS and STAR-RIS, covering architectural designs, signal models, and enabling techniques. Their study highlights the importance of protocol design and resource management for spectral efficiency, network coverage, and energy utilization. Wu et al.~\cite{Wu2024IEEE} consolidate the rapidly expanding literature on intelligent surfaces, covering not only traditional passive reflecting RIS but also advanced architectures such as active RIS, STAR-RIS, integrated reflection-refraction designs, and holographic beamforming surfaces capable of fine-grained electromagnetic wave manipulation \cite{wu2023ArXiv}. Complementing these foundations, Chopra et al.~\cite{chopra2025ris} examine RIS-assisted AISAC from a system-level perspective, analyzing energy-efficiency trade-offs, security vulnerabilities, spectrum utilization constraints, and hardware impairments. Tariq et al.~\cite{Tariq2025ACM} identify promising directions that integrate RIS with deep learning, digital twins, mobile edge computing, software-defined networking, UAVs, NOMA, and AISAC. 

A large body of work exploits RISs to jointly optimize sensing and communication. Luo et al.~\cite{Luo2022TVT} investigate joint beamforming for RIS-assisted AISAC, coordinating base-station precoding with RIS phase-shift optimization within a unified framework. Xing et al.~\cite{Xing2023TC} extend this idea by jointly optimizing active beamforming at the transmitter and passive phase-shift control at the RIS. Their algorithms address the inherent trade-off between sensing accuracy and communication quality under practical constraints, and demonstrate substantial gains over conventional RIS designs.
Zuo et al.~\cite{Zuo2023TVT} integrate RIS with NOMA, combining the spectral-efficiency advantages of NOMA with the channel reconfiguration capability of RIS to enhance sensing accuracy and communication performance while balancing competing objectives through efficient power allocation and beamforming. Wang et al.~\cite{Wang2023IOTJ} integrate RIS with backscatter communication to develop energy-efficient AISAC for large-scale IoT, jointly optimizing RIS phase configurations and waveform design to improve sensing precision and throughput under stringent power constraints. Gan et al.~\cite{Gan2024JSTSP} propose a double-RIS-aided scheme in which sensing pilots are superimposed onto data symbols rather than using separate pilot resources, significantly reducing overhead.

Several works address security and robustness, which are critical for practical RIS-assisted AISAC. Salem et al.~\cite{Salem2022TVT} study secure RIS-assisted AISAC in multi-input single-output (MISO) systems, showing that RISs can enhance sensing accuracy and communication security by controlling signal propagation and amplification. Yao et al.~\cite{Yao2025TVT} study hybrid RIS-aided AISAC in V2X networks, proposing a DRL approach that optimizes RIS configurations and beamforming to improve both sensing performance and communication security against eavesdroppers. Robustness under imperfect CSI is addressed in~\cite{Xu2024TVT}, whose transmission design explicitly accounts for channel estimation errors so that sensing and communication performance is maintained even when precise CSI is unavailable.
As network conditions, sensing targets, and user demands change, repeated re-optimization incurs significant computational overhead and signaling latency. Such designs also lack autonomous decision-making, which makes them difficult to scale to large, heterogeneous AISAC deployments that require real-time coordination among communication, sensing, and reconfigurable elements. 

\begin{figure}[t]
\begin{center}
    \includegraphics[width=0.6\columnwidth]{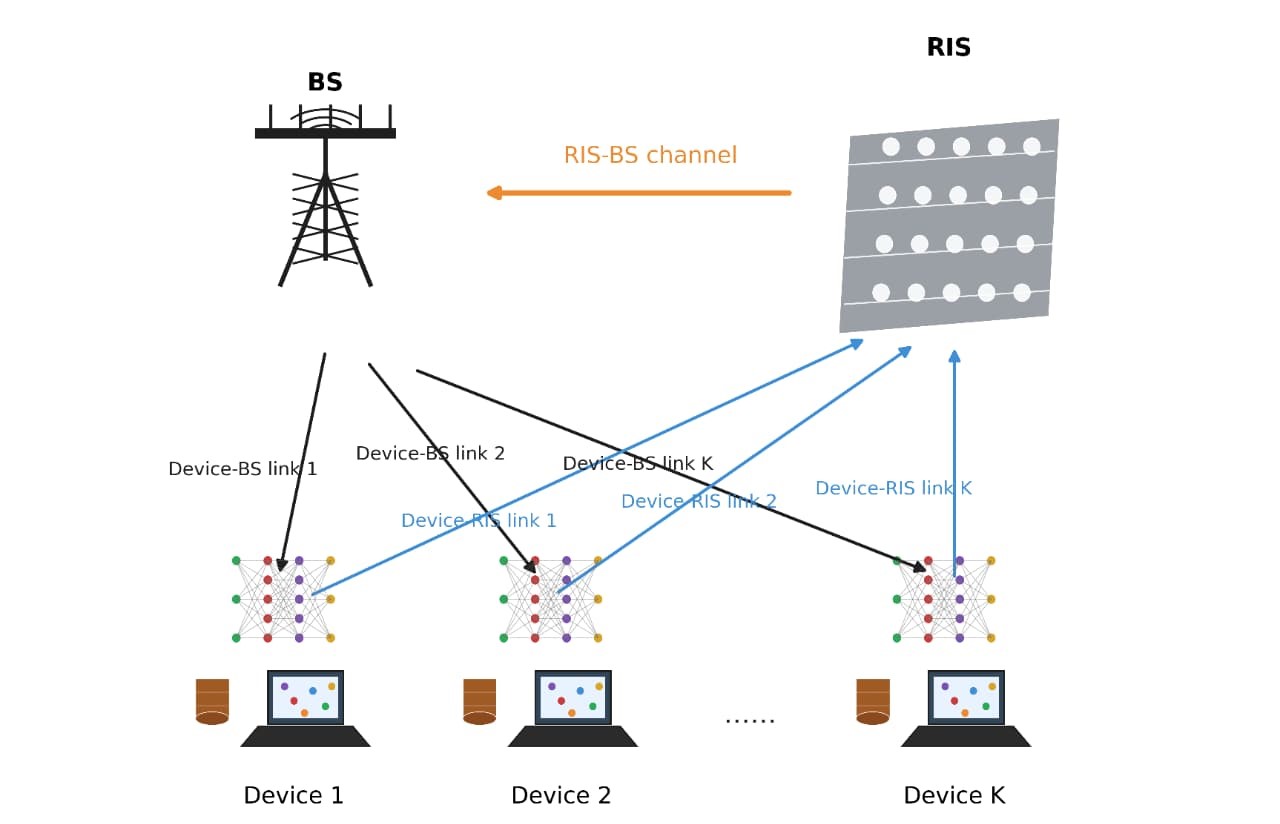}
\end{center}
\vspace{-1mm}
\caption{An Illustration of the considered wireless FL in an RIS-assisted NOMA System 
\cite{Sun2025TC}.}


\vspace{-2mm}
\label{fig:11}
\end{figure}

Advanced surface architectures expand the design space and call for new AISAC solutions. Zhou et al.~\cite{Zhou2025TGCN} investigate near-field extremely large-scale STAR-RIS, exploiting near-field propagation characteristics to provide highly directional communication links while achieving high-resolution sensing and localization. Zhu et al.~\cite{Zhu2025TC} introduce a reconfigurable holographic surface (RHS) as a unified platform for beamforming, environmental sensing, and data transmission within a single hardware platform, providing enhanced flexibility and hardware efficiency. Han et al.~\cite{Han2025TWC} explore multi-functional RIS for 6G AISAC, showing that a single surface can jointly enhance signal reflection, sensing accuracy, and communication reliability. Abdelaziz et al.~\cite{Abdelaziz2025TC} investigate integrated cooperative sensing and communication in RIS-enabled full-duplex cell-free MIMO systems, combining RIS with full-duplex operation and a cell-free architecture to improve both sensing accuracy and throughput while mitigating interference.

\begin{figure}[h]
\begin{center}
    \includegraphics[width=0.50\columnwidth]{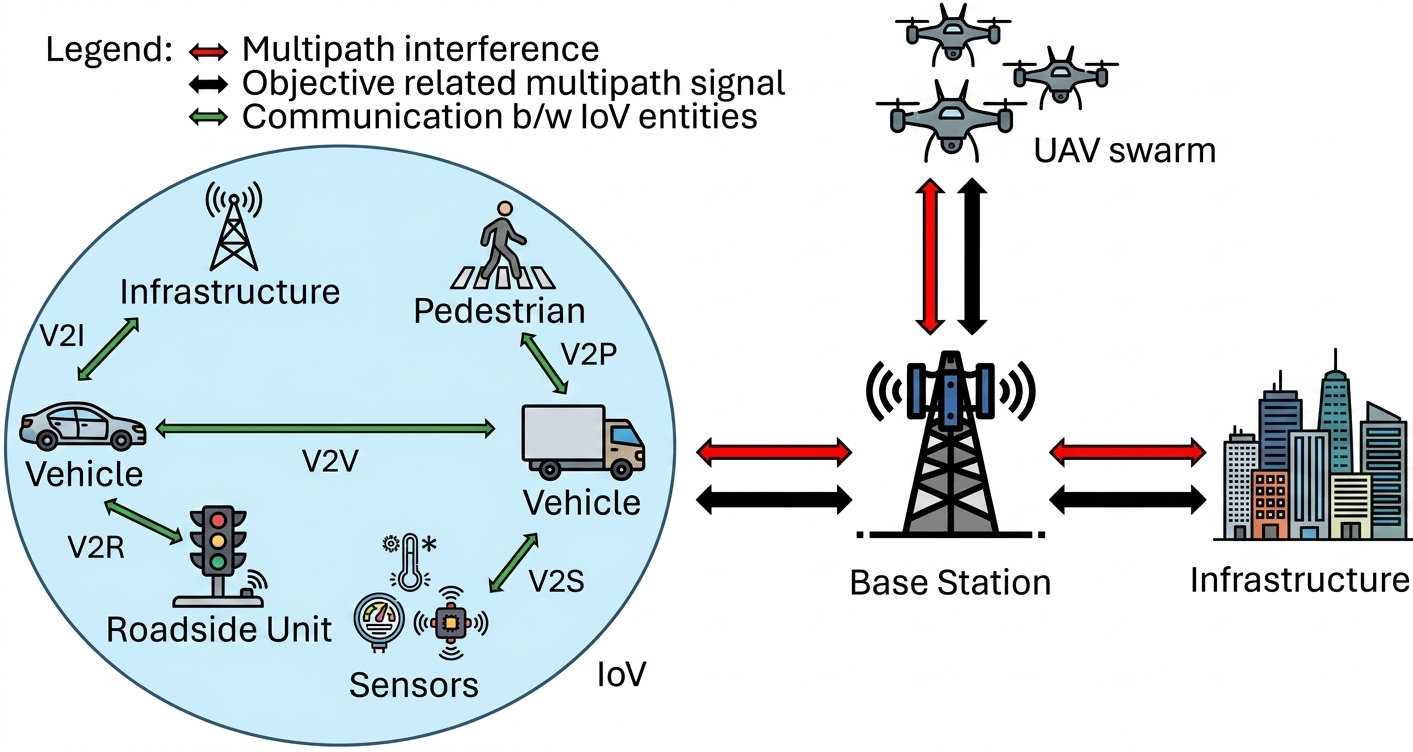}
\end{center}
\caption{ Overview of AISAC-enabled Low Altitude Network 
\cite{telecomlow}.}

\label{fig:aisac_lan}
\end{figure}

These architectural advances enable a range of emerging AISAC applications for next-generation intelligent environments. Ni et al.~\cite{Ni2025IoTM} apply RIS to the Internet of Robotic Things (IoRT), jointly optimizing transceiver beamforming, robot trajectories, and RIS reflection coefficients via multi-agent DRL to improve link quality, sensing accuracy, and energy utilization. Wu et al.~\cite{wu2024IOTM} introduce a RIS-assisted integrated sensing and backscatter communication system, using the RIS as a helper or low-power transceiver to overcome energy constraints and enable non-line-of-sight sensing. Guo et al.~\cite{Guo2025WC} build a RIS-enabled system that jointly supports communication, high-precision localization, and environmental mapping in complex vertical scenarios with three-dimensional structures, height variations, and indoor-outdoor transitions. Ma et al.~\cite{Ma2023JSTSP} leverage RIS to unify environmental sensing and high-quality communication for extended reality (XR), enhancing user experience through improved situational awareness, reliable connectivity, and low latency. In vehicular networks, Yang et al.~\cite{Yang2025IoTj} propose RIS-aided AISAC for Internet of Vehicles (IoV) systems, integrating communication, sensing, and edge-assisted computation under a joint beamforming design that satisfies vehicular quality-of-service requirements.

\subsection{Mobile Observation Platforms and Embodied Action Spaces}
\label{sec:physical_substrate:mobile}

Low-altitude networks are designed for operations in airspace below roughly 3 km altitude \cite{wu2025low}. An AISAC-enabled low-altitude network, illustrated in Fig.~\ref{fig:aisac_lan}, is an interconnected intelligent framework that unifies sensing, communication, and computing capabilities, using cellular mobile networks, the IoT, and cloud computing infrastructures~\cite{telecomlow, wang2025safeguarding}. As established earlier, RISs are widely considered across many domains, including aerial platforms and vehicular networks, to optimize drone and vehicular communications \cite{Han2025SJ}. It is essential to assess their current progress and fully unlock their potential in emerging communication scenarios.

Han et al.~\cite{Han2025SJ} review the status quo of intelligent vehicles and mobile communications, examine the key technologies and multi-domain applications of RISs, and outline existing challenges, recent advances, and future directions. Their study organizes each RIS application domain, such as vehicular or marine communications, around parameters such as the type of RIS, link direction, CSI, objectives, and optimization methods. Complementing this, the survey in~\cite{ahmed2025ArXiv} provides a comprehensive overview of UAV-based AISAC systems. It summarizes core concepts, recent technical advances, and key performance trade-offs, and highlights their potential to improve spectrum efficiency, reliability, latency, and energy usage. In particular,~\cite{ahmed2025ArXiv} focuses on beam tracking, channel estimation, and throughput, as well as trade-offs among weighted sum rates, sensing, delay, security, and energy efficiency.

\subsubsection{Existing Implementation of Agentic AI}
A first group of works already embeds AI within UAV-based AISAC systems. In \cite{Yang2021ISAC}, an actor-critic-based asynchronous FL algorithm is proposed to reduce overall system cost. Its device selection mechanism chooses a subset of devices with higher communication and computational capacity for FL model aggregation, thereby reducing processing time while minimizing accuracy loss. To reduce latency in UAV-enabled FL with AISAC, an algorithm based on block coordinate descent and SCA techniques is employed in \cite{Shaon2024CSCN}. It requires only four iterations to reach optimal latency, which is then maintained thereafter. The latency decreases significantly as the maximum frequency and transmit power of UAVs and base stations (BS) increase.
Other works refine which UAVs participate in learning. In \cite{Cheriguene2023PIMRC}, a UAV participant selection strategy, DEEPS, chooses the best FL node in each sub-region based on a dataset similarity score and energy profile.
In \cite{Ye2025TWC}, the authors apply DRL to optimize beamforming and UAV trajectories by modeling the problem as a Markov decision process.
\subsubsection{Opportunities for Agentic AI Integration}
A second group of works does not yet use agentic AI, but is well positioned for it. Cheng et al. \cite{Cheng2025TC} maximize the average UAV sum-rate during flight while satisfying sensing, power, and trajectory constraints, by leveraging alternating optimization, successive convex approximation, and semi-definite relaxation. 
Zhou et al. \cite{zhou2025ArXiv} present a MIMO-OFDM-based AISAC framework that operates without reserving specific time-frequency resources for sensing. This avoids overhead while improving both communication and sensing capabilities relative to traditional approaches. In~\cite{wang2025ArXiv}, a coordinated beamforming scheme integrates the optimization of dual-functional base station (DFBS) precoding and RIS passive beamforming, enhancing sum-rate while meeting sensing, power, and RIS constraints in an RIS-assisted AISAC system. 
None of these studies implements an agentic architecture; each optimizes a predefined FL or trajectory objective. Their optimization techniques for countering FL latency, participant selection, and energy constraints are nonetheless directly reusable as executable tools that a higher-level mission agent could invoke, which is what positions them for agentic integration rather than requiring them to be redesigned from scratch.

\begin{table*}[t]
\small
\centering
\caption{Representative physical-layer and infrastructure mechanisms in the Agentic AI--ISAC loop.}
\label{tab:physical_isac_summary}
\renewcommand{\arraystretch}{1.10}
\setlength{\tabcolsep}{4pt}
\begin{tabularx}{\textwidth}{
p{0.08\textwidth}
p{0.15\textwidth}
X
X
p{0.20\textwidth}}
\toprule
\textbf{Refs.} &
\textbf{Mechanism} &
\textbf{ISAC contribution} &
\textbf{Agentic role} &
\textbf{Main gap} \\
\midrule

\cite{Wu2024CM,zhu2023springer,Hu2025TWC} &
Task-oriented and edge-assisted ISAC &
Links sensing and communication resources with learning. &
Provides task-relevant observations and resource states. &
Predefined objectives and limited decomposition. \\

\cite{Yang2024TWC,Wen2024TWC,Li2023TWC} &
Joint ISAC optimisation &
Jointly controls beamforming, power, and feature processing. &
Provides executable allocation and computation tools. &
Limited semantic context, reasoning, and reflection. \\

\cite{Dou2023TVT,shang2025ArXiv} &
Scheduling adaptive beam control &
Adapts beamforming scheduling with transmission duration. &
Acts as a fast-timescale controller. &
Fixed rewards; little memory or tool-level reasoning. \\

\cite{Luo2022TVT,Xing2023TC,Zuo2023TVT,Wang2023IOTJ} &
RIS-assisted physical execution &
Controls beams, RIS phases, access, and backscatter links. &
Expands the agent's physical action space. &
Mostly centralised and goal-unaware. \\

\cite{Zhou2025TGCN,Zhu2025TC,Han2025TWC,Abdelaziz2025TC} &
Advanced programmable surfaces &
Supports coverage, localisation, and sensing through STAR-RIS. &
Enables programmable propagation and coverage control. &
High-dimensional control and limited validation. \\

\cite{Ni2025IoTM,Guo2025WC,Yang2025IoTj} &
Mobile and embodied ISAC platforms &
Integrates sensing, communication, mobility, and edge. &
Links network actions with physical movement. &
Incomplete safety-aware planning and verification. \\

\bottomrule
\end{tabularx}
\end{table*}
\subsection{Synthesis and Agentic Gap}
The studies reviewed above provide the physical measurements, radio configurations, and mobility variables required for closed-loop autonomy. Table~\ref{tab:physical_isac_summary} consolidates these mechanisms along with their role in the agentic loop and their remaining gap. As revealed, a complete agentic system must additionally translate mission-level goals into constraints, select among these tools, predict the consequences of alternative actions, and verify the outcome after execution.

\section{From ISAC Measurements to Contextual and Semantic Intelligence}
\label{sec:contextual_intelligence}

Physical observations become useful to an autonomous agent only after they are transformed into task-relevant context. This section reviews work that connects RF and network measurements with multimodal perception, semantic inference, mapping, and shared representations. These studies are mainly \emph{AI-centric} or \emph{interface-centric}: ISAC supplies measurements, while AI constructs the contextual state used by downstream reasoning and planning.

\subsection{Multimodal and Cross-Domain Contextualization}

\begin{figure}[t]
\begin{center}
    \includegraphics[width=0.6\columnwidth]{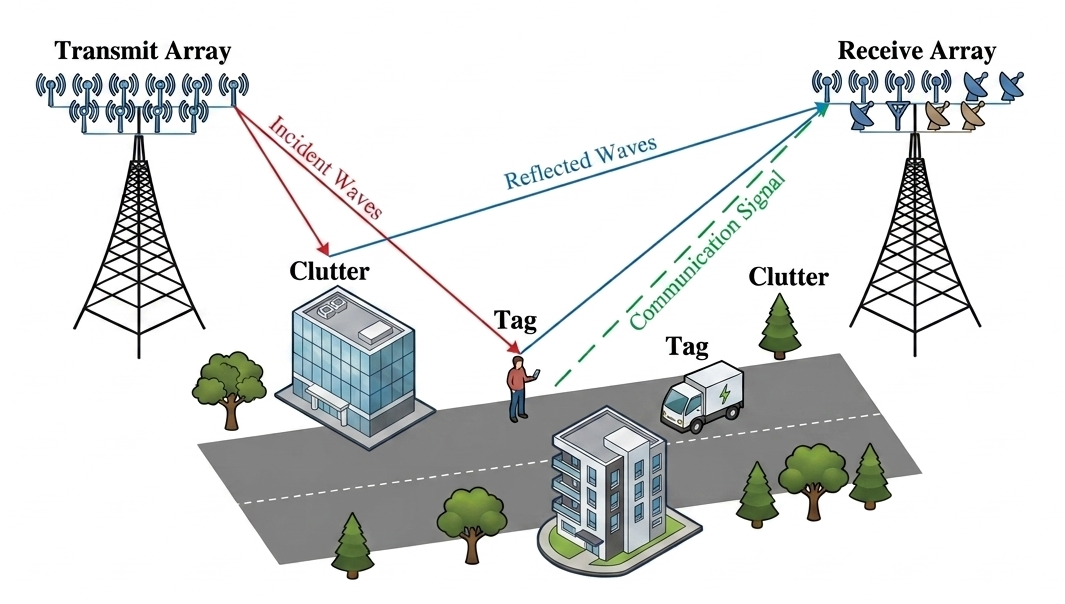}
\end{center}
\vspace{-1mm}

\caption{\small{Backscatter communication enables AISAC under stringent device power constraints \cite{Fang2025TNSE}.}}

\vspace{-2mm}
\label{fig:9}
\end{figure}

Efficient coordination, however, still assumes a relatively homogeneous sensing substrate. AISAC also expands beyond traditional single-waveform, single-sensor assumptions toward multi-modal, heterogeneous, and cross-domain architectures. This expansion is driven by the need to operate reliably in complex environments and to support diverse edge intelligence tasks. A representative direction enhances the physical-layer sensing-communication substrate, for example by incorporating ultra-low-power backscatter links and movable-antenna configurations.
The survey and framework in \cite{Fang2025TNSE} position backscatter communication as an effective solution to stringent device power constraints, such as sub-mW operation (Fig.~\ref{fig:9}). It also highlights that accurate tag localization in cluttered, short-range environments cannot be fully addressed by classical time of arrival (ToA), angle of arrival (AoA), or direction of arrival (DoA) techniques alone. 

A second expansion axis concerns heterogeneous sensing modalities and semantics, where RF radar alone becomes insufficient for robust perception in dynamic environments. The LLM-driven M-AISAC vision in \cite{Liu2025IoTM} frames multi-modal AISAC as the fusion of radar with LiDAR, cameras (RGB-D), and GPS. It is motivated by the complementary strengths and failure modes of non-RF sensors, such as weather and illumination sensitivity, and by the limited expressiveness of radar-only perception. The authors of~\cite{Liu2025IoTM} argue that LLMs provide a unifying reasoning and representation layer for context-aware fusion, adaptive sensing control, and cross-modal transfer. This pushes AISAC toward semantic consistency and task-generalizable operation across IoT scenarios. 

A third direction focuses on edge-intelligence-driven AISAC, where system design can be jointly optimized around data generation, transmission, and inference. The AISAC-enabled edge intelligence frameworks in \cite{Zhang2022ICC} and \cite{Zhang2023TGCN} describe a fundamental paradigm shift. Instead of a sequential sense-then-transmit pipeline, sensing and communication are tightly integrated, allowing wireless signals to concurrently support data acquisition and uploading. This joint operation can reduce end-to-end latency in centralized edge learning systems. The achievable improvement, however, depends heavily on the relative time scales of data generation and transmission. 

Recent studies have extended AISAC beyond traditional radar–communication integration to heterogeneous sensing, distributed learning, and cross-layer optimization. Huang et al.~\cite{Huang2025TCCN} optimize RIS phase shifts, image resolution, and transmit power to integrate radar sensing, image sensing, and edge inference within a unified AISAC framework. FL has also been adopted to address privacy, communication overhead, and scalability in heterogeneous wireless systems, including mmWave massive MIMO~\cite{Ardianto2024IoTj} and V2X networks with adaptive multiple access~\cite{Chen2024Network}. Furthermore, Fu et al.~\cite{Fu2025TVT} extend AISAC to multi-UAV over-the-air FL by jointly designing sensing, communication, and distributed computation as an integrated optimization problem.


\subsection{Cross-Domain Tracking, Mapping, and XR Context}

RIS-assisted sensing and learning have been investigated for cross-domain and data-scarce scenarios, enabling knowledge transfer across heterogeneous environments. X2Track~\cite{Hu2024JSAC} uses RIS-enabled channel reconfiguration to extract informative multi-band features and transfer tracking knowledge from well-labeled source domains to sparsely labeled target domains. Although it is not agentic, its RIS control and domain-adaptation mechanisms could be incorporated as executable tools for adaptive sensing, reflection configuration, and real-time learning coordination.

\begin{table*}[t]
\small
\centering
\caption{Representative approaches for transforming ISAC observations into contextual and semantic intelligence.}
\label{tab:contextual_intelligence_summary}
\renewcommand{\arraystretch}{1.10}
\setlength{\tabcolsep}{4pt}
\begin{tabularx}{\textwidth}{
p{0.08\textwidth}
p{0.15\textwidth}
X
X
p{0.20\textwidth}}
\toprule
\textbf{Refs.} &
\textbf{Function} &
\textbf{ISAC input} &
\textbf{AI contribution} &
\textbf{Main gap} \\
\midrule

\cite{Wu2024CM,zhu2023springer,Wen2024TWC} &
Task-oriented feature construction &
RF observations, sensing quality, states, and edge constraints. &
Learns task-relevant features related to accuracy and latency. &
Task-specific, limited generalisation and memory. \\

\cite{Tian2025WC,wei2025large} &
Multimodal contextualisation &
RF/radar data with camera, LiDAR, GPS, or telemetry. &
Aligns modalities and derives scene, object, and context. &
Hallucination, reliability, and real-time cost. \\

\cite{Xiong2024IoTj,Guo2025WC} &
Tracking, localisation, and mapping &
Multi-band or RIS-assisted measurements. &
Constructs spatial context and transferable representations. &
Limited goal reasoning and action selection. \\

\cite{Zhang2023TIV} &
Cooperative world modelling &
Distributed vehicular sensing and communication data. &
Builds a federated SLAM with reduced raw-data exchange. &
Map consistency and asynchronous updates. \\

\cite{Zhou2025WC} &
Multi-agent reconstruction &
Distributed visual or three-dimensional observations. &
Uses specialised agents for reconstructing context. &
Weak coupling with physical ISAC and feedback. \\

\cite{Ma2023JSTSP,Yang2025IoTj} &
Application-specific semantics &
Environmental connectivity information for real-time services. &
Produces situational awareness and service-level context. &
Application-specific context is poorly reused. \\

\bottomrule
\end{tabularx}
\end{table*}

\subsection{Vehicular Mapping and Cooperative World Representation}
\label{sec:contextual_intelligence:vehi}

Thanks to their high manoeuvrability and adaptable deployment, UAVs are suited to provide AISAC services to vehicles operating in dynamic environments \cite{liu2025uav}. The works discussed below are not framed as AISAC, but employ AI for efficient vehicular networks.
Zhang et al. \cite{Zhang2023TIV} replace costly phased-array antennas with affordable RHS antennas and enable collaborative sensing through FL, which helps autonomous driving systems significantly improve simultaneous localization and mapping (SLAM) performance. The authors show that the issues arising from the unique radiation characteristics of RHS, together with the complexities of federated information exchange, can be addressed through a multi-vehicle SLAM protocol.
Zhang et al. \cite{Zhang2023JSAC} propose a rate-splitting multiple access-based IoV framework that integrates FEEL-supported downlink and platoon control. The downlink and platoon-control latency of FEEL is reduced iteratively, using SCA for optimal communication and MPC for platoon control.
Although these works do not explicitly integrate ISAC and agentic AI, they use AI techniques and address the challenges AI introduces to improve vehicular network performance.

\subsection{Distributed Reconstruction and Specialized Perception Agents}

Beyond UAVs and vehicles, AISAC has promising applications across 6G and beyond \cite{dogru2025panama}. The studies below discuss AISAC deployment in future networks without specifically targeting UAVs or vehicles, yet their techniques apply readily to either or both. Dogru et al. \cite{dogru2025panama} employ MARL to optimize multi-agent path finding, together with centralized training and decentralized execution under asynchronous architectures, to enable efficient autonomous task execution. The approach delivers a higher success rate with reduced solution length, shows enhanced resilience, and suffers fewer blackout events than benchmarks. Similarly, Liang et al. \cite{Liang2025TNSE} optimize energy efficiency for 3D object reconstruction and transmission by leveraging a multi-agent mixture of experts. 
The scheme highlights the benefits of combining decentralized coordination with specialized expertise.

\subsection{Synthesis and Agentic Gap}
These studies, summarized in Table~\ref{tab:contextual_intelligence_summary}, show how physical measurements can be converted into object-, scene-, and task-level representations. However, contextualization alone is not equivalent to agency. Most existing systems do not explicitly connect the constructed context to persistent memory, uncertainty-aware goal reasoning, multi-step planning, or verified ISAC tool use. These missing interfaces motivate the predictive and planning functions reviewed next.

\section{Prediction, Goal Reasoning, and Agentic Planning}
\label{sec:reasoning_planning}

This section reviews the cognitive and orchestration mechanisms that turn contextual states into future-oriented decisions, from adaptive resource-control policies to intent interpretation, generative solution construction, digital-twin support, and AI-native network management. We distinguish policy-centric controllers from stronger agentic components that incorporate goals, context, tools, and feedback.

\subsection{Intent Interpretation, Generative Intelligence, and AI-Native Control}

\begin{figure}[t]
\begin{center}
    \includegraphics[width=0.50\columnwidth]{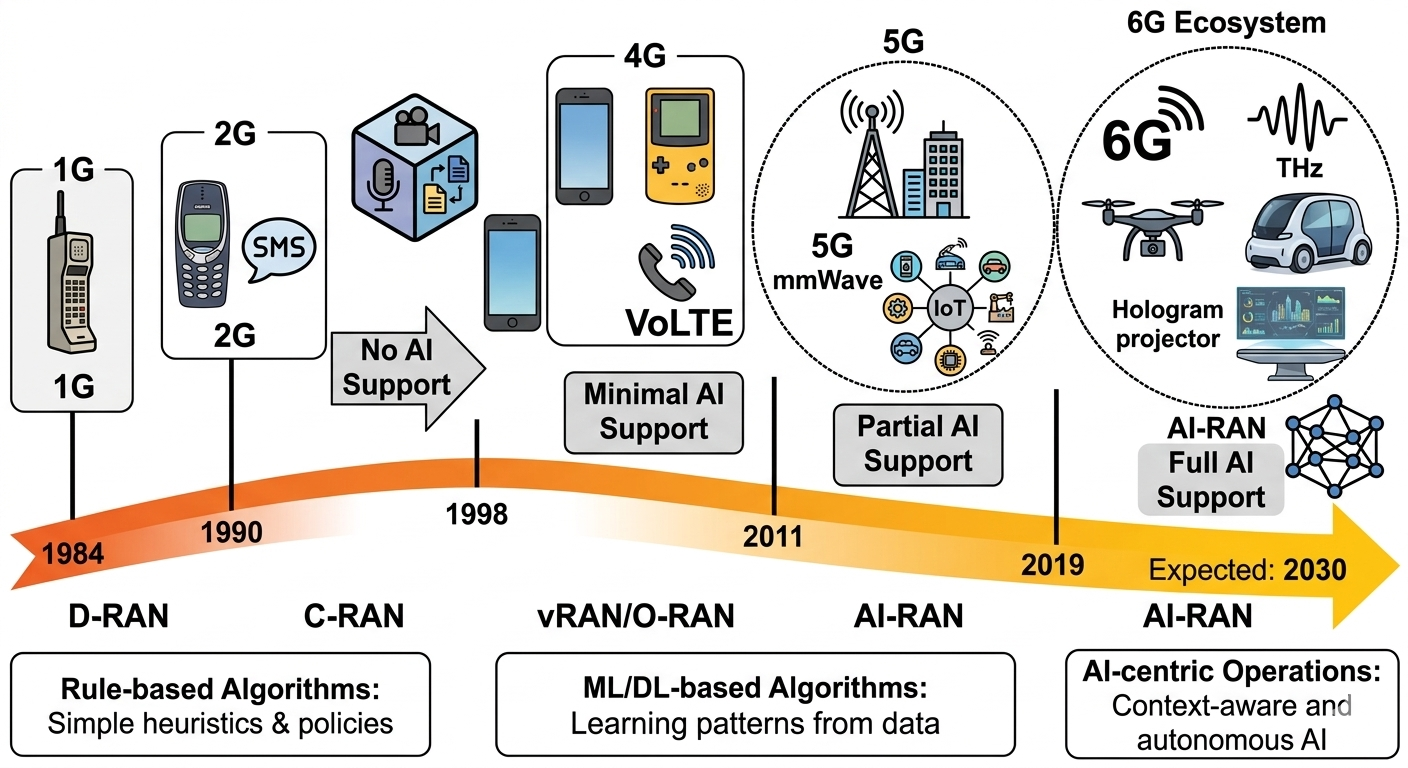}
\end{center}
\vspace{-1mm}

\caption{\small{Evolution of mobile networks and RAN architectures from the perspective of AI adoption \cite{rathakrishnan2025towards}.}}

\vspace{-2mm}
\label{fig:10}
\end{figure}

Multi-modal and cross-domain advances demand a cognitive layer that can reason over them. AI-native 6G reframes AISAC from task-specific optimization blocks into autonomy-centric architecture with cognition embedded in network control loops, since future environments are too dynamic for rule-based or narrowly trained controllers, pushing radio access network (RAN) toward self-evolving, self-healing operation \cite{rathakrishnan2025towards} (Fig.~\ref{fig:10}). Sensing and communication become jointly orchestrated actuators configured proactively based on context, predicted dynamics, and intents.





AI-driven RAN consolidates orchestration across layers and timescales, integrating digital twins, intelligent surfaces, and generative intelligence for autonomous decision-making \cite{rathakrishnan2025towards}. Wireless large AI models extend foundation-model paradigm to wireless systems, addressing isolated models' poor cross-task generalization; positioned as versatile network brains, they both optimize the network and are trained and deployed through it via edge intelligence and collaborative learning \cite{zhu2025wireless}, supporting unified reasoning over heterogeneous telemetry rather than brittle task-by-task tuning.

Generative AI enables scalable solution synthesis for both one-shot decisions and sequential control modeled as Markov decision processes \cite{yang2025frontiers}, but struggles with strict constraint satisfaction, symbolic reasoning, and mathematical reliability when data is scarce and latency-critical. This suggests an agentic division of roles, where generative models propose candidate actions and agentic controllers enforce feasibility and safe execution.

O-RAN offers a substrate for this division via modular control across near-real-time and non-real-time controllers, though practical orchestration challenges remain: conflict resolution among independently developed xApps/rApps, synchronized context sharing, and translating operator intent into executable policies \cite{salmi2025ai}. Combining DRL-driven conflict management, context synchronization, LLM-based intent translation, and RIS-integrated closed-loop control shows how multi-agent orchestration can bridge sensing, communication, and computation in one control fabric, with LLMs as semantic controllers that interpret unstructured intents and collaborate with solvers for coupled optimization \cite{wei2025large}. Prompt engineering -- in-context learning, chain-of-thought reasoning, iterative self-refinement -- offers a resource-efficient path to this without expensive retraining \cite{Zhou2025WC}.

Sustaining such autonomy requires efficient adaptation under heterogeneous devices and non-IID data; knowledge-sharing distributed fine-tuning, e.g., aggregating heterogeneous LoRA updates with parameter-efficient procedures, adapts large pretrained models while cutting communication and computation overhead \cite{Wang2025TNSE}.


\subsection{Learning-Enabled RIS Control and Distributed Policy Optimization}
\label{sec:reasoning_planning:learn}

Integrating learning with RIS opens directions for privacy-preserving, distributed intelligence, marking an early shift toward agentic AISAC. Das et al.~\cite{Das2024OJCS} survey FL combined with reconfigurable intelligent metasurfaces, showing RIS-controlled propagation can enhance reliability while FL preserves privacy; decentralized FL aligns naturally with agentic principles, since RIS controllers, edge devices, and network nodes can be modeled as autonomous agents learning and acting on local observations.
Several works realize this alignment. Pala et al.~\cite{Pala2025TCCN, Pala2024GLOBECOM} propose FL frameworks for joint beamforming and sensing over RIS-enhanced satellite-terrestrial networks, combining distributed optimization, adaptive RIS-assisted beamforming, and privacy-preserving collaborative learning without sharing raw data, improving scalability and robustness through decentralized aggregation. A related study~\cite{Pala2025TWC} shows decentralized FL over hybrid satellite-terrestrial networks improves privacy, scalability, and adaptability, with open challenges in aggregation under non-IID data and dynamic topologies. Sun et al.~\cite{Sun2025TC} study RIS-enhanced wireless FL under imperfect aggregation (Fig.~\ref{fig:11}), showing RISs can mitigate aggregation errors by shaping wireless channels during learning, and develop joint RIS/FL optimization improving accuracy and efficiency.

Reinforcement learning also supports autonomous control: Wu et al.~\cite{Wu2024ICC} propose federated RL coordinating multiple dual-star RISs in DFRC-enabled multi-base-station AISAC, learning policies without exchanging raw data. Other paradigms include over-the-air personalized FL via RIS \cite{shi2024ArXiv}, RIS-empowered topology control for decentralized FL in urban air mobility \cite{Xiong2024IoTj}, and latency-aware FL integrating STAR-RIS with wireless power transfer \cite{Alishahi2025IoTj}.

These FL-RIS frameworks fit the AISAC paradigm, letting network nodes, RIS elements, and sensing modules function as autonomous agents that perceive, learn, and act while optimizing beamforming, resource allocation, and sensing under privacy and robustness constraints.

\subsection{Toward Agent-Controlled Programmable Environments}

Static, centrally optimized RIS control is increasingly inadequate for dynamic AISAC, and agentic AI offers a way forward through synergy: each agent contributes specialized capabilities, planning, perception, execution, evaluation, while adapting to others, enabling emergent behavior beyond isolated automation.
Modeling RIS controllers, transmitters, and sensing modules as cooperative agents lets the system monitor the environment, detect anomalies, and reconfigure in real time. For STAR-RIS, dedicated agents can control transmission/reflection modes and phase shifts while coordinating with communication and sensing agents under channel variation and mobility, enabling dynamic mode-switching and proactive defense against interference and adversarial behavior, rather than static, pre-optimized strategies.

Reprogrammability lets agents update goals, policies, or tools without rebuilding the system, incorporating new objectives or technologies with minimal friction for robust, future-proof deployments. Representative steps include the ISRiD framework~\cite{Ahmad2024IoTj}, integrating AISAC, RIS, and DDPG-based RL for Internet-of-Everything applications that learn control policies from feedback, and computation-aware designs such as the beamforming/resource-allocation framework of~\cite{Bian2025IoTj} and the radar-detection optimization of~\cite{Xiao2024TVT}.

\subsection{Planning versus Adaptive Control}
The reviewed DRL, MARL, optimization, and model-predictive approaches provide adaptive control, but most assume predefined rewards, objectives, and action spaces: the RAN/O-RAN orchestration layers~\cite{rathakrishnan2025towards,salmi2025ai} select among a fixed set of xApps rather than composing new ones from intent; the RIS-FL frameworks~\cite{Pala2025TCCN,Pala2024GLOBECOM,Sun2025TC,Wu2024ICC} learn beamforming and aggregation under a single fixed objective; and the agent-controlled RIS designs~\cite{Ahmad2024IoTj,Bian2025IoTj,Xiao2024TVT} adapt phase configurations within a predetermined reward structure. In our stage labeling (Section~\ref{sec:framework:literature_alignment}), these are therefore best classified as R (reasoning and prediction) or bounded P (planning), since they optimize within a closed objective and action space rather than composing a plan from an open-ended goal -- planning components or callable controllers, not complete agents, consistent with the closed-world/open-world distinction in Section~\ref{sec:framework:agent}.

Stronger Agentic AI--ISAC systems should instead support intent translation, goal decomposition, multi-timescale prediction, tool selection, constraint verification, and replanning. Only LLM-based and generative-reasoning approaches above~\cite{wei2025large,yang2025frontiers,Zhou2025WC} begin to exhibit this, and even they need an external mechanism to enforce feasibility once a plan is generated, since neither the generative model nor the underlying RIS/FL controller checks this itself. Table~\ref{tab:planning_summary} positions each mechanism along this spectrum, from single-objective solvers to the more open-ended, though still incompletely verified, agent-oriented architectures.

\begin{table*}[t]
\small
\centering
\caption{Representative AI and control mechanisms for prediction, reasoning, and Agentic AI--ISAC planning.}
\label{tab:planning_summary}
\renewcommand{\arraystretch}{1.10}
\setlength{\tabcolsep}{4pt}
\begin{tabularx}{\textwidth}{
p{0.08\textwidth}
p{0.15\textwidth}
X
X
p{0.20\textwidth}}
\toprule
\textbf{Refs.} &
\textbf{Mechanism} &
\textbf{Decision capability} &
\textbf{ISAC connection} &
\textbf{Main gap} \\
\midrule

\cite{rathakrishnan2025towards,tang2025towards} &
Agent-oriented architecture &
Interprets intent, workflows, and specialised agents. &
Maps high-level goals to sensing and communication. &
Limited physical-layer integration and verification. \\

\cite{shi2024ArXiv,wei2025large} &
Generative and LLM reasoning &
Interprets context, and transfers knowledge across tasks. &
Formulates optimisation problems or resource control. &
Requires trusted, latency-aware tools. \\

\cite{Yang2024TWC,shang2025ArXiv,Ni2025IoTM} &
DRL/MARL adaptive control &
Learns state-dependent power, beam, RIS, and trajectories. &
Controls ISAC resources in vehicular and robotic settings. &
Fixed rewards, goals, and memory dominate. \\

\cite{Sun2025TC,Pala2025TCCN,Pala2024GLOBECOM} &
Federated policy optimisation &
Learns distributed control policies without centralising data. &
Supports decentralised RIS, beamforming, and resource. &
Limited negotiation and coordination. \\

\cite{Luo2022TVT,Xing2023TC,Xu2024TVT} &
Optimisation and robust solvers &
Computes beamforming and RIS settings under constraints. &
Acts as an executable tool for a higher-level agent. &
No explicit reasoning, planning, or tool switching. \\

\cite{Zhang2023JSAC,Wang2024JSAC} &
AI-native/O-RAN interfaces &
Coordinates applications, policies, and closed-loop control. &
Translates agent decisions into network actions. &
Tools conflict, runtime accountability. \\

\bottomrule
\end{tabularx}
\end{table*}

\subsection{Synthesis and Agentic Gap}

The literature progresses from static optimization to learning-enabled adaptation and AI-native orchestration (Table~\ref{tab:planning_summary}), but evidence for fully closed-loop Agentic AI remains limited. The most credible near-term architecture is hierarchical: a cognitive agent interprets goals and selects tools, while fast optimizers, policies, and physical controllers execute bounded actions at radio and mobility time scales.

\section{Distributed and Multi-Agent ISAC Execution}
\label{sec:distributed_execution}

AISAC systems are naturally distributed across base stations, edge servers, RIS controllers, UAVs, vehicles, robots, and user devices. This section asks a question those sections leave open: how do these distributed components become a coordinated multi-agent system? We organize the discussion around distributed learning and synthesis, coordination topology, and role assignment, before turning to the security considerations and agentic implications.

\subsection{Synthesis of Distributed Learning and Embodied Coordination}

The mechanisms that make Agentic AI--ISAC distributed have already been introduced earlier in this survey. Section~\ref{sec:reasoning_planning:learn} reviewed how FL integrates with RIS control to support decentralized, privacy-preserving policy optimization across ISAC entities, ranging from FL-metasurface frameworks~\cite{Das2024OJCS} and satellite-terrestrial joint beamforming and sensing~\cite{Pala2024GLOBECOM, Pala2025TCCN, Pala2025TWC}, to imperfect-aggregation-aware RIS--FL~\cite{Sun2025TC} and federated reinforcement learning for multi-RIS DFRC systems~\cite{Wu2024ICC}, with further extensions to over-the-air personalized FL~\cite{shi2024ArXiv}, topology control for urban air mobility~\cite{Xiong2024IoTj}, and STAR-RIS-integrated wireless power transfer~\cite{Alishahi2025IoTj}. Sections~\ref{sec:physical_substrate:mobile} and~\ref{sec:contextual_intelligence:vehi} further showed how UAV-enabled AISAC systems apply FL and DRL for device/UAV participant selection, latency reduction, and joint trajectory-beamforming design~\cite{Yang2021ISAC,Shaon2024CSCN,Cheriguene2023PIMRC,Ye2025TWC,Cheng2025TC,zhou2025ArXiv,wang2025ArXiv}, and how vehicular and robotic deployments combine cooperative sensing, federated SLAM, platoon control, multi-agent path finding, and specialized multi-agent reconstruction~\cite{Zhang2023TIV,Zhang2023JSAC,dogru2025panama,Liang2025TNSE}.

Rather than restating these results, Table~\ref{tab:distributed_summary} re-organizes them specifically along the distributed-coordination dimension of the AISAC loop: for each mechanism, it identifies the distributed contribution, its role in enabling closed-loop operation, and the coordination-specific limitation that remains open. Two points emerge from this synthesis that are not visible when the same works are read individually. First, FL-based and DRL-based coordination mechanisms are consistently better developed on the learning side than on the agentic side: they optimize shared policies or models, but rarely include explicit role negotiation, capability discovery, or mission-level task allocation among nodes. Second, UAV and vehicular studies increasingly treat participant selection, trajectory planning, and resource scheduling as joint problems, which is a necessary but not sufficient step toward treating UAVs, vehicles, and RIS controllers as autonomous agents with persistent goals and memory, as required by the agentic maturity levels defined in Section~\ref{sec:framework:literature_alignment}.

\subsection{Coordination Topologies in Distributed AISAC}
\label{sec:topologies}

The mechanisms reviewed in Sections~\ref{sec:physical_substrate:mobile}, \ref{sec:contextual_intelligence:vehi}, and~\ref{sec:reasoning_planning:learn} implicitly adopt one of three coordination topologies, which we make explicit here because the choice of topology strongly constrains how agentic a system can become.

\textit{Centralized aggregation topologies}, exemplified by conventional FL-based RIS control~\cite{Das2024OJCS,shi2024ArXiv} and UAV-enabled FL~\cite{Yang2021ISAC,Shaon2024CSCN,Cheriguene2023PIMRC}, route all model updates through a single server or base station. This topology simplifies convergence analysis and enables straightforward Byzantine-robust aggregation, since a single aggregator can inspect and filter all incoming updates before combining them, but it reintroduces a single point of coordination that sits with the agentic principle of distributed,
goal-directed decision-making: the aggregator decides how local contributions are combined.

\textit{Hierarchical topologies}, such as the federated reinforcement learning scheme for multi-dual-STAR-RIS coordination~\cite{Wu2024ICC} and the satellite-terrestrial FL frameworks~\cite{Pala2024GLOBECOM,Pala2025TCCN,Pala2025TWC}, introduce an intermediate coordination layer (e.g., a regional controller or gateway) between edge nodes and a global model. This mirrors the three-loop hierarchy proposed in Section~\ref{sec:framework:hier} (fast physical control, medium-timescale orchestration, slow cognitive planning), and is the topology most compatible with Level-3/Level-4 agentic maturity, since intermediate agents can, in principle, perform local planning and tool selection before escalating to a mission agent.

\textit{Peer-to-peer and swarm topologies} appear in UAV trajectory and beamforming coordination~\cite{Cheng2025TC,zhou2025ArXiv,wang2025ArXiv,Ye2025TWC} and in multi-agent path finding~\cite{dogru2025panama}, where nodes exchange information directly without a fixed aggregator. These topologies are the most naturally agentic in structure, since decision-making is genuinely distributed, but the reviewed literature rarely equips peer nodes with persistent memory, negotiation protocols, or explicit goal representations; most peer-to-peer schemes still optimize a shared, predefined reward rather than reconciling potentially conflicting local objectives.

None of the reviewed topologies is inherently agentic; the topology only determines where coordination decisions could, in principle, be made. Whether these decisions are genuinely produced through reasoning, negotiation, and adaptation rather than by a fixed protocol or static aggregation rule is examined more closely below, where we revisit what it means for a distributed component to be truly agentic.

\subsection{Role Assignment and Heterogeneous Capability Matching}
\label{sec:role_assignment}

A second, implicit, dimension across the literature is how nodes are assigned functional roles. Three patterns recur.

\textit{Capability-based selection} chooses participants according to a measurable property, such as computational or communication capacity~\cite{Yang2021ISAC}, dataset similarity and energy profile~\cite{Cheriguene2023PIMRC}, or link quality and trajectory feasibility~\cite{Ye2025TWC,Cheng2025TC}. This pattern improves system-level efficiency but treats role assignment as a one-shot filtering step rather than a continuing negotiation; selected nodes do not subsequently reassess their suitability as conditions change.

\textit{Specialization-based assignment}, exemplified by the multi-agent mixture-of-experts framework for 3D reconstruction~\cite{Liang2025TNSE} and by RIS-enabled Internet-of-Robotic-Things coordination~\cite{Ni2025IoTM}, assigns different agents to different sub-tasks (e.g., sensing versus reconstruction versus communication scheduling) based on fixed functional roles defined at design time. This is consistent with the agent-role taxonomy introduced in Section~\ref{sec:framework:agent} (mission, context, prediction, sensing/communication, computation, infrastructure, and security agents), but the reviewed works instantiate only one or two of these roles at a time, rather than a full multi-role team.

\textit{Adaptive team formation}, in which the set of participating nodes and their roles change dynamically in response to failures or mobility, is the least developed pattern in the literature. UAV-assisted hierarchical FL under dynamic smart IoT conditions~\cite{yang2025ArXiv} is one of the few works that reconfigures participation in response to node loss, but reconfiguration follows a predefined rule rather than an agent-level decision informed by predicted mission impact.

These patterns show that role assignment in current AISAC systems is largely static or rule-based. Genuine capability discovery and negotiation, where nodes advertise their state and an orchestrating agent (or the nodes themselves) decide role allocation based on mission context, remains absent and is an open direction; see Section~\ref{sec:future_agentic}.
\begin{table*}[t]
\small
\centering
\caption{Representative distributed and multi-agent mechanisms in Agentic AI-enabled ISAC.}
\label{tab:distributed_summary}
\renewcommand{\arraystretch}{1.10}
\setlength{\tabcolsep}{4pt}
\begin{tabularx}{\textwidth}{
p{0.09\textwidth}
p{0.14\textwidth}
X
X
p{0.20\textwidth}}
\toprule
\textbf{Refs.} &
\textbf{Mechanism} &
\textbf{Distributed contribution} &
\textbf{Closed-loop role} &
\textbf{Main gap} \\
\midrule

\cite{Liu2021TWC,Hu2025TWC,Ouyang2024SJ} &
Federated over-the-air learning &
Aggregation under resource overhead constraints. &
Supports shared learning and adaptation across ISAC nodes. &
Goals, roles, and plan-level coordination. \\

\cite{Pala2025TCCN,Pala2024GLOBECOM,Sun2025TC} &
Federated RL for RIS control &
Learns decentralised policies for programmable surfaces. &
Links distributed policy learning with physical ISAC. &
Policies conflict, convergence, and safety. \\

\cite{Liu2025IoTM,Mu2023IWCMC,Ye2025TWC} &
UAV participation and coordination &
Selects participants and allocates system resources. &
Supports dynamic team formation and edge collaboration. &
Limited task semantics and coalition reasoning. \\

\cite{yang2025ArXiv,Zhang2023TIV} &
Vehicular learning and control &
Combines perception, federated mapping and control. &
Connects shared context with trajectory and platoon actions. &
Latency, safety, and heterogeneous sensor alignment. \\

\cite{Ni2025IoTM,Zhou2025WC} &
Robotic and specialised-agents &
Coordinates mobility, RIS control, and expert assignment. &
Enables embodied execution and specialised role allocation. &
Limited mission generalisation and interoperability. \\

\cite{Wang2025TNSE,Fang2025TNSE} &
Resilient topology adaptation &
Reassigns participants and collaboration under failures. &
Supports feedback-driven replanning continuity. &
Lacks layer-specific mission-level verification. \\

\bottomrule
\end{tabularx}
\end{table*}

\subsection{Synthesis and Agentic Gap}
\label{sec:agentic-interpretation}
FL clients, RIS controllers, and mobile nodes should not automatically be labeled as agents. A distributed component becomes agentic only when it possesses an explicit objective, contextual state, action capability, and feedback mechanism. The reviewed literature provides many of these building blocks, including distributed learning, adaptive policies, topology control, trajectory optimization, and cooperative perception, but integrating them into a coherent multi-agent architecture remains an open challenge. The security and resilience implications of this distributed structure -- including the attack surface it creates and the failure-recovery mechanisms it requires -- are treated separately in Section~\ref{sec:trust_resilience}, since they concern the feedback and resilience stage of the closed loop rather than distributed execution itself.

\section{Feedback, Trust, and Resilience Across the Agentic Loop}
\label{sec:trust_resilience}


An understanding of any technology is incomplete without a clear view of its security threats and its resilience against them. Systems that integrate ISAC and agentic AI are susceptible to threats targeting either technology. As discussed in~\cite{Naeem2023OJCS}, RIS-enhanced AISAC offers substantial gains in spectral efficiency, target detection, and interference suppression. Its LOS-dominated air-ground channels, however, make it vulnerable to eavesdropping, jamming, and attacks from malicious UAVs. The authors of \cite{Naeem2023OJCS} examine the security and privacy challenges that arise when RISs are integrated with 6G technologies such as AISAC, discussing attack types, and specific target areas. As noted earlier, FL is widely used to address the storage, computation, and energy limitations of AISAC. This reliance, in turn, exposes such systems to FL-specific attacks, such as inference and poisoning attacks, on top of AISAC-related threats \cite{sagar2023poisoning}. The agentic variants of these attacks, such as memory poisoning and tool misuse, are an additional concern \cite{john2025owasp}.

\begin{table*}[t]
\centering
\caption{Trust, security, and resilience mechanisms across the six-stage Agentic AI--ISAC closed loop.}
\label{tab:resilience_aisac}
\footnotesize
\setlength{\tabcolsep}{3.2pt}
\renewcommand{\arraystretch}{1.10}
\begin{tabularx}{\textwidth}{
>{\raggedright\arraybackslash}p{1.75cm}
>{\raggedright\arraybackslash}p{2.05cm}
>{\raggedright\arraybackslash}X
>{\raggedright\arraybackslash}X
>{\raggedright\arraybackslash}p{2.35cm}}
\toprule
\textbf{Stage} &
\textbf{Protected asset} &
\textbf{Threats / failures} &
\textbf{Mechanisms and works} &
\textbf{Main gap} \\
\midrule

\textbf{Observation (O)} &
Waveforms, RF data, CSI, localisation, and provenance &
Spoofing, jamming, eavesdropping, waveform manipulation, and corrupted sensing &
Secure beamforming, secrecy control, interference mitigation, and provenance protection
\cite{Wang2024JSAC,Meng2025TVT,Mu2023IWCMC,Naeem2023OJCS} &
Limited tracing of corrupted observations through later decisions. \\
\midrule

\textbf{Contextualisation (C)} &
Multimodal fusion, semantic state, and shared context &
Inference attacks, semantic leakage, poisoned features, and cross-modal inconsistency &
Privacy-preserving learning, semantic protection, anomaly detection, and consistency checking
\cite{Paul2025IoTj,tang2025towards,sagar2023poisoning} &
Limited physical grounding and uncertainty-aware context verification. \\
\midrule

\textbf{Reasoning and prediction (R)} &
Memory, knowledge, predictive models, and reasoning traces &
Memory poisoning, prompt injection, manipulated knowledge, and biased prediction &
Provenance-aware memory, guarded retrieval, input validation, and uncertainty monitoring
\cite{john2025owasp,Rai2024GCCIT} &
Limited auditability and recovery from compromised memory. \\
\midrule

\textbf{Planning and orchestration (P)} &
Goals, plans, tool selection, and control policies &
Goal hijacking, unsafe plans, malicious instructions, and tool misuse &
Policy constraints, least-privilege access, safety checks, logging, and rollback
\cite{john2025owasp,al2025defense} &
Feasibility checks rarely verify intent or plan-level safety. \\
\midrule

\textbf{Execution and collaboration (E)} &
FL aggregation, agent messages, UAV/RIS control, and actuation &
Byzantine or Sybil agents, poisoned updates, interception, and node compromise &
Participant filtering, robust aggregation, clustered or blockchain-assisted FL, and covert communication
\cite{Akram2025JSTAEORS,Hafeez2023CAMAD,yang2025ArXiv,Tong2024TWC} &
Limited dynamic trust, role reassignment, and execution verification. \\
\midrule

\textbf{Feedback and resilience (F)} &
Telemetry, diagnosis, topology adaptation, and recovery &
Node or link failure, poisoned feedback, cascading errors, and energy depletion &
Predictive maintenance, resilient FL, adaptive pairing, anomaly detection, and reconfiguration
\cite{Tian2025WC,zhu2022resilient,yang2025ArXiv,Rai2024GCCIT,al2025defense} &
Recovery remains layer-specific and rarely closes the mission loop. \\
\bottomrule
\end{tabularx}
\end{table*}

Table~\ref{tab:resilience_aisac} summarizes representative resilience mechanisms for AISAC systems. Existing studies address resilience from multiple perspectives, including Byzantine-robust aggregation, adversarial defense, failure recovery, and cross-layer protection. Significant progress has been made on individual threats, yet most solutions target a specific attack surface, such as FL, wireless communication, or UAV networking.

\subsection{Byzantine Robustness and Aggregation}
Byzantine-robust aggregation addresses the risk that a subset of participating nodes returns corrupted, stale, or maliciously crafted updates that would otherwise bias the shared model. In post-disaster UAV monitoring, Akram et al.~\cite{Akram2025JSTAEORS} propose a conditional FL approach for privacy-preserving spatial crowdsourcing, in which drone contributions are accepted only when they satisfy trust and relevance conditions, reducing the influence of unreliable or adversarial participants without requiring raw data to be shared. Hafeez et al.~\cite{Hafeez2023CAMAD} take a structural approach instead, proposing a blockchain-enabled clustered and scalable FL framework for UAV networks: clustering isolates the effect of a compromised UAV to its local group, while the blockchain ledger provides a tamper-evident record of model updates that supports auditable, verifiable aggregation at scale. Yang et al.~\cite{yang2025ArXiv} address robustness from the participation side, optimizing energy, latency, and resilience jointly in an adaptive UAV-assisted hierarchical FL framework for dynamic smart IoT; by reconfiguring which UAVs relay and aggregate updates as conditions change, the scheme limits the aggregation-time exposure of any single unreliable or adversarial node. These works show that Byzantine robustness in AISAC is pursued through complementary levers, conditional participant filtering, structural clustering with tamper-evident logging, and adaptive relay selection.

\subsection{Adversarial Defenses}
A second group of studies defends AISAC against eavesdropping, jamming, and inference threats that target the physical and semantic layers directly. Wang et al.~\cite{Wang2024JSAC} use an active aerial RIS to integrate sensing and positioning with secure communication, exploiting the RIS's active amplification and phase control to steer signal energy toward legitimate receivers while suppressing leakage toward potential eavesdroppers. Meng et al.~\cite{Meng2025TVT} pursue a closely related goal from the waveform design side, formulating ISAC transmission under explicit communication security-rate constraints. Paul and Singh~\cite{Paul2025IoTj} extend adversarial robustness into the decision-making layer, proposing a privacy-preserved, distributed multi-agent federated DRL framework for 6G-IoV that couples trustworthy AI principles with dynamic electric-vehicle charging and task-offloading decisions. Tong et al.~\cite{Tong2024TWC} address a different threat model, covert communication, by designing UAV-assisted FL over mmWave massive MIMO in which the very existence of the FL traffic is concealed from a warden. Tang et al.~\cite{tang2025towards} target the semantic layer, studying secure semantic communications in the presence of intelligent eavesdroppers capable of interpreting transmitted meaning, which requires defenses that operate on the semantic representation. These defenses span the physical layer (RIS-assisted secrecy, security-rate-constrained sensing), the communication layer (covert transmission), and the semantic and decision layers (multi-agent DRL privacy, semantic-level secrecy). Adversarial defense in AISAC cannot be reduced to a single physical-layer security technique.

\subsection{Link and Node Failure Recovery}
A third perspective addresses resilience to link and node failures rather than adversarial manipulation. Tian et al.~\cite{Tian2025WC} examine trustworthy 6G-powered industrial IoT for resilient intelligent manufacturing, where predictive maintenance and trust-aware scheduling are used to anticipate node degradation before it disrupts the sensing-communication-computation pipeline, rather than only reacting after a failure occurs. Zhu et al.~\cite{zhu2022resilient} instead target the learning layer, proposing resilient and communication-efficient learning for heterogeneous federated systems in which clients differ in computational capability, connectivity, and availability; their design tolerates intermittent or dropped client participation without requiring every node to remain continuously reachable. Yang et al.~\cite{yang2025ArXiv}'s adaptive UAV-assisted hierarchical FL framework contributes directly to failure recovery: by dynamically re-pairing devices with UAVs as connectivity or energy conditions change, the framework maintains FL convergence and mission continuity even as individual UAV-device links become temporarily unavailable. Across these three studies, failure recovery in AISAC is addressed through prediction (anticipating degradation before it occurs), tolerance (learning algorithms that degrade gracefully under dropout), and adaptive reconfiguration (dynamically reassigning roles as the network topology changes), rather than through a single fault-recovery mechanism applied at one layer.

\subsection{Cross-layer Resilience}

A fourth line of work develops cross-layer resilience mechanisms rather than protecting individual subsystems in isolation. Rai et al.~\cite{Rai2024GCCIT} jointly apply machine learning and network analytics to improve 6G safety, privacy, and resource efficiency, while Mu et al.~\cite{Mu2023IWCMC} address secure sensing-data transmission and sharing in distributed ISAC. Al-Karaki~\cite{al2025defense} further advocates a defense-in-depth architecture that combines complementary protections across the physical, network, and application layers. Such multilayered resilience is essential because poisoned updates, spoofed sensing data, and adversarial waveforms can propagate throughout the sensing–communication–computation pipeline.

\subsection{Synthesis and Agentic Gap}
The reviewed mechanisms protect individual components, including physical-layer secrecy, robust aggregation, model privacy, predictive maintenance, and topology recovery. A complete resilient agentic architecture must connect these mechanisms through end-to-end integrity checks: observation provenance, contextual consistency, safe plan verification, controlled tool access, anomaly detection, task migration, and verified fallback. Robustness without diagnosis and replanning is insufficient for mission-level autonomy.

\begin{figure*}[t]
\begin{center}
 \includegraphics[width=0.8\columnwidth]{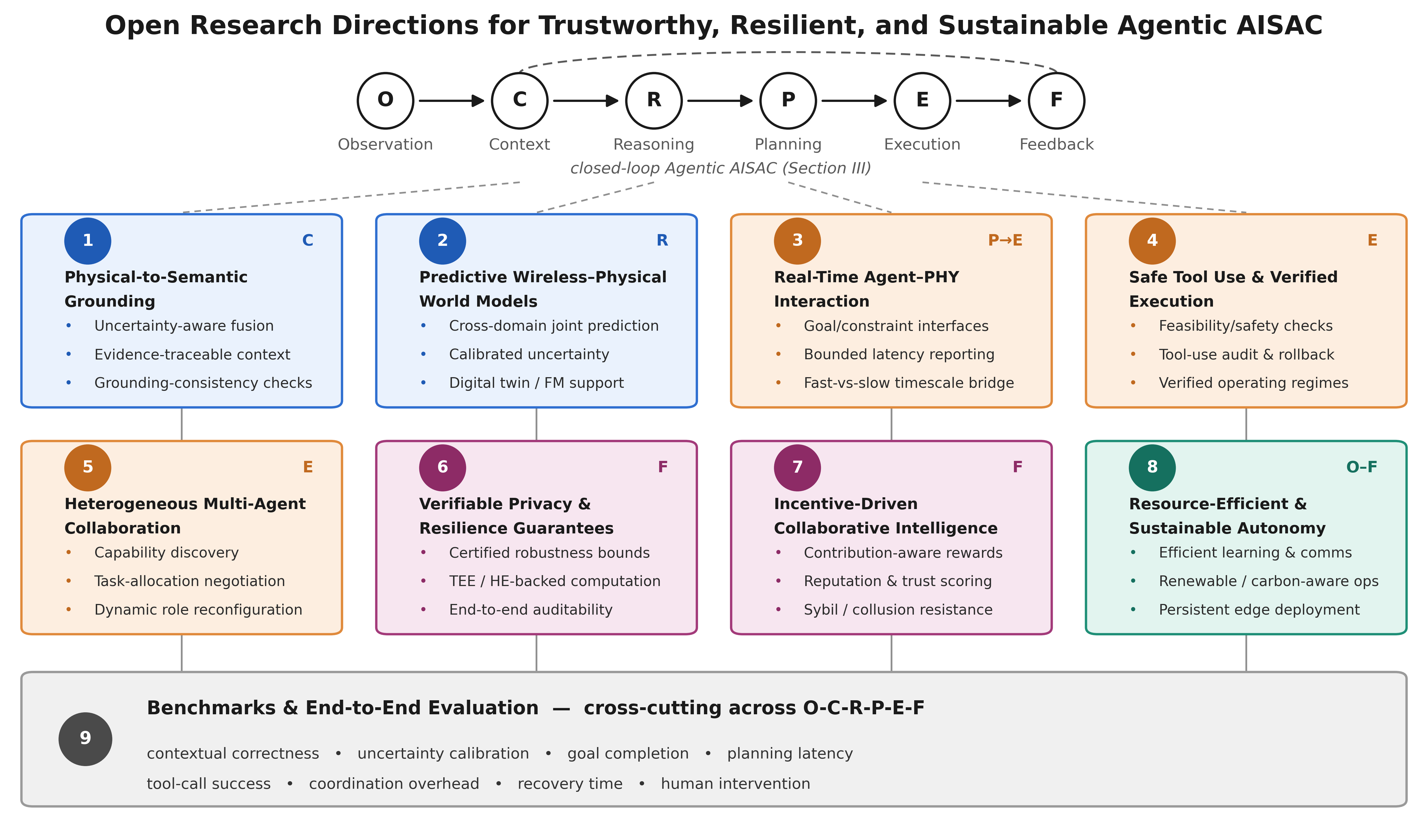}
\end{center}
\caption{Research directions for advancing privacy-preserving and resilient solutions in future AISAC systems.}
\label{fig:future}
\end{figure*}

\section{Open Challenges and Future Research Directions}
\label{sec:future_agentic}
Sections~\ref{sec:physical_substrate}--\ref{sec:trust_resilience} show substantial progress within individual loop stages, but isolated bodies of work, not a coherent agenda, address grounding, real-time interaction, tool use, coordination, sustainability, and evaluation. This section organizes eight remaining gaps, summarized in Fig.~\ref{fig:future}: physical-to-semantic grounding, predictive world models, real-time agent--PHY interaction, safe tool use, heterogeneous multi-agent collaboration, verifiable privacy and resilience guarantees, incentive-driven collaboration, and resource-efficient autonomy -- each tied to the stage labels (O, C, R, P, E, F) and maturity levels (L0--L4) from Section~\ref{sec:framework}, since these gaps are connective tissue between levels rather than new problems.

\vspace{2mm}
\noindent \textbf{Physical-to-semantic grounding.} Section~\ref{sec:contextual_intelligence}'s contextualization-stage (C) studies map observations to object-, scene-, and network-level semantics end-to-end, with no account of \emph{why} a label follows from a given reading. This weakens what an agent needs most: tracing a decision back to its evidence and recognizing when that evidence is insufficient or manipulated. Grounding must become a design target, not a by-product. Promising ingredients: uncertainty-aware fusion propagating sensing confidence (CRB, SNR) into semantics; knowledge-augmented context tied to verifiable measurements; and consistency checks flagging when $c^{sem}_t$ is unsupported by $o^{phy}_t$. Without these, agentic ISAC inherits multimodal models' reliability limits and exposes a physical attack surface.

\vspace{2mm}
\noindent \textbf{Predictive wireless--physical world models.} Agents need joint, calibrated models of objects, mobility, channels, interference, traffic, load, energy, and threats over a horizon $\hat{s}_{t+1:t+H}$, so planning anticipates rather than reacts. Work under the reasoning-and-prediction stage (R, Section~\ref{sec:reasoning_planning}) typically predicts one domain in isolation without propagating uncertainty forward. Two gaps follow: world models must couple physical, wireless, and computational state in one representation rather than separate outputs; and they must return calibrated uncertainty, not point estimates, since a planner (P) can only trade off objectives if it trusts each prediction accordingly. Digital-twin and foundation-model approaches are natural candidates, but their cost must fit edge latency and energy budgets.

\vspace{2mm}
\noindent \textbf{Real-time agent--PHY interaction.} High-level reasoning runs at hundreds of milliseconds to seconds; beam tracking and safety-critical control need slot- or millisecond-level responses. This mismatch blocks the loop between planning (P) and execution (E). The hierarchy in Section~\ref{sec:framework} (mission agent $\rightarrow$ orchestrator $\rightarrow$ schedulers $\rightarrow$ physical controllers) offers a structural answer but leaves open: how a slow agent expresses goals for a fast controller without re-deriving them each step; how the physical layer reports anomalies upward without flooding the agent with telemetry; and how responsibility is apportioned when a controller must deviate from a plan for safety. Verified, bounded-latency interfaces are needed, not unstructured message passing.

\vspace{2mm}
\noindent \textbf{Safe tool use and verified execution.} Section~\ref{sec:framework:agent} framed ISAC optimizers, RIS controllers, and digital twins as tools an agent invokes -- convenient, but it hides a safety gap: a wrong tool choice, misconfiguration, or out-of-regime call is invisible in the tool's own metrics. Memory poisoning, prompt injection, and tool misuse (Section~\ref{sec:trust_resilience}) apply directly here. Closing this requires feasibility checks between plan and controller, tool-use logging for audit and rollback, and verification of when a tool's output is trustworthy; otherwise, L3--L4 systems gain tool-oriented coordination alongside an unexamined attack surface.

\vspace{2mm}
\noindent \textbf{Heterogeneous multi-agent collaboration.} Section~\ref{sec:role_assignment} found roles are mostly assigned through one-shot selection or fixed specialization, with little genuine negotiation among heterogeneous agents. Treating the mission, context, prediction, sensing/communication, computation, infrastructure, and security roles (Section~\ref{sec:framework:agent}) as an interacting team requires agents to advertise state, agree on task allocation under partial observability, and reconfigure as nodes join, fail, or move. This ties to coordination topology (Section~\ref{sec:topologies}): hierarchical topologies suit such negotiation best, mediating without a single aggregator or full peer-to-peer consensus, but concrete protocols remain undeveloped.

\vspace{2mm}
\noindent \textbf{Verifiable privacy and resilience guarantees.} Section~\ref{sec:trust_resilience} found most defenses target a single attack surface, not the system as a whole. Above them sits a further gap: agentic AISAC offers little \emph{verifiable} assurance that a deployed system behaves as intended, versus per-component guarantees. Differential-privacy accounting, certified robustness bounds, trusted-execution or homomorphic-encryption computation, and formal policy verification are promising, but combining them into one auditable, end-to-end guarantee remains open.

\begin{table*}[t]
\centering
\caption{Coverage of agentic-specific evaluation dimensions in representative AISAC studies reviewed in this survey.}
\label{tab:benchmark_gap}
\resizebox{\textwidth}{!}{
\begin{tabular}{p{2.6cm}p{3.8cm}p{2.5cm}p{3cm}p{4.5cm}}
\toprule
\textbf{Study category} &
\textbf{What is currently reported} &
\textbf{Representative works} &
\textbf{Agentic dimensions reported} &
\textbf{Agentic dimensions missing} \\
\midrule

Learning-driven ISAC / UAV control &
Comparison of convergence iterations, sum-rate, and communication throughput &
\cite{Cheng2025TC,zhou2025ArXiv,wang2025ArXiv,Ye2025TWC} &
Planning latency (partial, via convergence speed) &
Contextual correctness, goal completion, tool-call success, recovery time, human intervention \\

\midrule

FL participant selection / device scheduling &
Training accuracy vs. baseline, dataset-similarity or energy profile &
\cite{Cheriguene2023PIMRC} &
Coordination overhead (partial, via participant selection cost) &
Contextual correctness, uncertainty calibration, goal completion, tool-call success, recovery time \\
\midrule

Resilience and robustness mechanisms &
Attack-specific robustness (e.g., poisoning tolerance, aggregation accuracy) &
\cite{Akram2025JSTAEORS,Hafeez2023CAMAD,yang2025ArXiv} &
Recovery time (partial, via failure-recovery latency) &
Contextual correctness, goal completion, planning latency, tool-call success, reduced overhead \\
\midrule

Agent-oriented network architectures &
Qualitative case studies and architectural walk-throughs; no quantitative metric &
\cite{LI2025Engineering,coelho2025a4fn,wei2025large} &
None reported quantitatively &
All nine dimensions reported only descriptively, not measured \\

\midrule

Agentic resource allocation in ISAC &
Task-level performance (e.g., bandwidth prioritization outcome) under a single deployment scenario &
\cite{alshalwi2025toward} &
Goal completion (partial, single-task) &
Contextual correctness, uncertainty calibration, planning latency, tool-call success, reduced overhead, recovery time, human intervention \\

\bottomrule
\end{tabular}}
\end{table*}

\vspace{2mm}
\noindent \textbf{Incentive-driven collaborative intelligence.} Federated and multi-agent AISAC assumes vehicles, UAVs, RIS controllers, and edge devices will contribute data, computation, and updates to a shared objective. In practice, resource-constrained nodes have different owners, costs, and incentives, so participation and honesty cannot be assumed. Contribution-aware reward allocation, reputation and trust scoring, and ledger-based verification are needed to sustain collaboration under free-riding, while staying lightweight for edge nodes and resistant to collusion and Sybil attacks.

\vspace{2mm}
\noindent \textbf{Resource-efficient and sustainable autonomy.} Closed-loop operation at L3--L4 adds persistent cognitive and communication overhead -- memory, prediction, planning, coordination -- atop the physical ISAC and learning costs already surveyed, which resource-constrained UAVs, vehicles, and IoT devices cannot treat as secondary. Efficient learning (sparse/low-rank updates, distillation, early exit), communication (semantic, adaptive scheduling), and computation (offloading, task migration) must be co-designed with renewable, carbon-aware infrastructure so the whole pipeline is optimized jointly, not combined post hoc -- a precondition for persistent, not just demonstration-scale, deployment.

\vspace{2mm}
\noindent \textbf{Benchmarks and end-to-end evaluation.} Evaluation should move beyond physical-layer metrics (rate, Cram\'er--Rao bound, latency, accuracy) toward nine agentic dimensions: contextual correctness, uncertainty calibration, goal completion, planning latency, constraint violations, tool-call success, coordination overhead, recovery time, and human-intervention frequency, audited in Table~\ref{tab:benchmark_gap}. Even the strongest examples -- ISAC controllers reporting convergence and sum-rate~\cite{Cheng2025TC,zhou2025ArXiv,wang2025ArXiv,Ye2025TWC}, participant selection reporting accuracy~\cite{Cheriguene2023PIMRC}, resilience reporting attack robustness~\cite{Akram2025JSTAEORS,Hafeez2023CAMAD,yang2025ArXiv}, agent-oriented architectures reported qualitatively~\cite{LI2025Engineering,coelho2025a4fn,wei2025large}, and the one agentic ISAC resource-allocation study~\cite{alshalwi2025toward} -- report at most one or two dimensions, none covering contextual correctness, tool-call success, or human intervention. 

Closing this gap needs reproducible benchmarks exposing both physical-layer dynamics and agent-level tasks in one environment (e.g., a testbed with ground truth for the sensed environment, mission goal, and resource budget) so stage-by-stage contribution (O, C, R, P, E, F) and overall maturity (Section~\ref{sec:framework:literature_alignment}) can be assessed consistently. Community suites, as in embodied AI and LLM agent evaluation, with standardized logging of tool calls, plans, and recovery, would enable comparison on common footing.

\section{Conclusion}
\label{sec:conclusion}
As future 6Gnetworks evolve toward autonomous, distributed intelligence, AISAC marks a significant paradigm shift. It moves away from conventional communication-centric architectures and toward deeply integrated, intent-driven sensing and communication ecosystems. By embedding agentic AI into ISAC, intelligent edge entities can collaboratively perform environmental perception, distributed learning, adaptive communication, and autonomous decision-making under dynamic and heterogeneous wireless conditions. To chart this paradigm, the survey reviewed the architectural evolution, enabling technologies, and emerging security challenges of AISAC. In particular, we highlighted how the tight coupling among sensing and communication opens new privacy leakage channels and resilience vulnerabilities. 
Building on this analysis, we identified open challenges spanning the full closed loop, including physical-to-semantic grounding, predictive wireless--physical world models, real-time agent--PHY interaction, safe tool use and verified execution, heterogeneous multi-agent collaboration, verifiable privacy and resilience guarantees, incentive-driven collaborative intelligence, and resource-efficient, sustainable autonomy, together with the reproducible benchmarks needed to evaluate them consistently. These directions point toward a trustworthy, autonomous, and sustainable foundation for AISAC in the 6G and Metaverse era.

\bibliographystyle{IEEEtran} 
\bibliography{references}

\end{document}